%% file: main.tex
\documentclass[sigconf]{acmart}
\usepackage{bm}
\usepackage{algorithm}
\usepackage{multirow}
\usepackage[noend]{algorithmic}
\usepackage{subfigure}
\usepackage{tikz}
\usepackage{url}
\usepackage{xspace}
\newcommand*{\circled}[1]{\lower.7ex\hbox{\tikz\draw (0pt, 0pt)circle (.5em) node {\makebox[1em][c]{\small #1}};}}

\usepackage{wrapfig}
\usepackage{subfigure}
\usepackage{multirow}
\usepackage{bm}
\usepackage{bbm}
\usepackage{color, soul}
\usepackage{subfigure}
\usepackage{graphicx}
\usepackage{listings}
\usepackage{xcolor}

\usepackage{makecell}
\usepackage{color, soul}
\usepackage{bm}
\usepackage{booktabs}

\usepackage{xcolor}
\newcommand{\red}[1]{\textcolor{red}{#1}}

\newcommand{\sys}{\textsc{ROD}\xspace}

\AtBeginDocument{%
  \providecommand\BibTeX{{%
    \normalfont B\kern-0.5em{\scshape i\kern-0.25em b}\kern-0.8em\TeX}}}

\copyrightyear{2021}
\acmYear{2021}
\setcopyright{acmcopyright}\acmConference[KDD '21]{Proceedings of the 27th ACM SIGKDD Conference on Knowledge Discovery and Data Mining}{August 14--18, 2021}{Virtual Event, Singapore}
\acmBooktitle{Proceedings of the 27th ACM SIGKDD Conference on Knowledge Discovery and Data Mining (KDD '21), August 14--18, 2021, Virtual Event, Singapore}
\acmPrice{15.00}
\acmDOI{10.1145/3447548.3467221}
\acmISBN{978-1-4503-8332-5/21/08}

\begin{document}
\fancyhead{}

\title{\sys: Reception-aware Online Distillation for Sparse Graphs}
\author{Wentao Zhang$^\dagger$, Yuezihan Jiang$^{\dagger\ddagger}$, Yang Li$^{\dagger}$, Zeang Sheng$^{\dagger}$, Yu Shen$^{\dagger}$, \\
Xupeng Miao$^{\dagger}$, 
Liang Wang$^{\ddagger}$, 
Zhi Yang$^{\dagger}$, 
Bin Cui$^{\dagger \S}$}
\affiliation{
 {{$^\dagger$}{School of EECS \& Key Laboratory of High Confidence Software Technologies, Peking University}}~~~~~
 {{$^{\S}$}{Center for Data Science, Peking University \& National Engineering Laboratory for Big Data Analysis and Applications}}~~~~~
$^\ddagger$Alibaba Group\country{}}
\affiliation{
$^\dagger$\{wentao.zhang, jiangyuezihan, liyang.cs, shengzeang18, shenyu, xupeng.miao, yangzhi, bin.cui\}@pku.edu.cn\\ $^\ddagger$\{jiangyuezihan@jyzh, liangbo.wl\}@alibaba-inc.com\country{}
}

\renewcommand{\shortauthors}{Wentao Zhang, et al.}

\begin{abstract}
Graph neural networks (GNNs) have been widely used in many graph-based tasks such as node classification, link prediction, and node clustering. However, GNNs gain their performance benefits mainly from performing the feature propagation and smoothing across the edges of the graph, thus requiring sufficient connectivity and label information for effective propagation. Unfortunately, many real-world networks are sparse in terms of both edges and labels, leading to sub-optimal performance of GNNs. Recent interest in this sparse problem has focused on the self-training approach, which expands supervised signals with pseudo labels. Nevertheless, the self-training approach inherently cannot realize the full potential of refining the learning performance on sparse graphs due to the unsatisfactory quality and quantity of pseudo labels.

In this paper, we propose ROD, a novel reception-aware online knowledge distillation approach for sparse graph learning. We design three supervision signals for ROD: multi-scale reception-aware graph knowledge, task-based supervision, and rich distilled knowledge, allowing online knowledge transfer in a peer-teaching manner. To extract knowledge concealed in the multi-scale reception fields, ROD explicitly requires individual student models to preserve different levels of locality information. For a given task, each student would predict based on its reception-scale knowledge, while simultaneously a strong teacher is established on-the-fly by combining multi-scale knowledge. Our approach has been extensively evaluated on 9 datasets and a variety of graph-based tasks, including node classification, link prediction, and node clustering. The result demonstrates that ROD achieves state-of-art performance and is more robust for the graph sparsity.
\end{abstract}




\maketitle



{\fontsize{8pt}{8pt} \selectfont
\textbf{ACM Reference Format:}\\
Wentao Zhang, Yuezihan Jiang, Yang Li, Zeang Sheng, Yu Shen, Xupeng Miao, Liang Wang, Zhi Yang, Bin Cui. 2021. ROD: Reception-aware Online Distillation for Sparse Graphs. In \textit{Proceedings of the 27th ACM SIGKDD Conference on Knowledge Discovery and Data Mining (KDD ’21), August 14–18, 2021, Virtual Event, Singapore. } ACM, New York, NY, USA, 11 pages. https://doi.org/10.1145/3447548.3467221 }

\input{1.intro}
\input{2.pre}

\input{3.method}
\input{4.experiment}

\section{Conclusion}
In this paper, we proposed ROD, the first online knowledge distillation framework supporting various types of tasks for sparse graphs. 
A key idea of ROD is that preserving and distilling the knowledge from different scales of reception fields to improve the graph learning performance.
To this end, \sys uses decoupled GNN student models to preserve knowledge from different levels of localities, and make complementary predictions based on students' own reception-scale knowledge. Each student gets its own supervision from specific tasks, reception-aware knowledge, and distilled multi-scale knowledge from the ensemble teacher. In this way, \sys effectively addresses the sparsity problem for graph learning.
We conduct extensive experiments to verify the effectiveness of \sys with 9 datasets on different tasks such as node classification, link prediction, and node clustering. The results showed \sys provided substantial improvement over the state-of-the-art competitors on various graph-based tasks, and additional high training efficiency.

\begin{acks}
This work is supported by the National Key Research and Development Program of China (No. 2018YFB1004403), NSFC (No. 61832001, 6197200),  and Beijing Academy of Artificial Intelligence (BAAI). Zhi Yang and Bin Cui are the corresponding authors.
\end{acks}

\bibliographystyle{ACM-Reference-Format}
\bibliography{my-reference}
\input{5.appendix}

\end{document}

%% file: 1.intro.tex
\vspace{-3mm}
\section{Introduction}
Graph neural networks (GNNs) generalize deep neural networks (DNNs) to graph-structured data, and they have achieved state-of-art performance in various tasks such as node classification~\cite{kipf2016semi,spinelli2020adaptive}, link prediction~\cite{kipf2016variational,wu2020garg}, and node clustering~\cite{zhang2019attributed,cui2020adaptive}.
GNNs primarily derive their benefits from performing feature propagation and
smoothing over graph neighborhoods, which captures the graph structure as well as features of individual nodes.
Unfortunately, most graphs exhibit sparsity characteristics~\cite{wang2019time} in terms of:
 (1) \underline{Edge sparsity}: real-world graphs that often have the skewed degree distribution, suffering from the severe sparsity problem, especially when the largest degree is small~\cite{kuramochi2005finding}.
(2) \underline{Label sparsity}: only a small set of nodes are labeled, especially when expert annotators are required~\cite{garcia2017few}.
Such graph sparsity inherently limits the effective information propagation and the performance of GNNs ~\cite{sun2020multi,cui2020adaptive}.

The most promising direction to address the sparse issues faced by GNNs is to adopt self-supervising strategies~\cite{baresi2010self,you2020does}. Traditional approaches along this direction adopt the self-training~\cite{zou2019confidence} method. For the node classification task, multi-stage self-supervised learning (M3S)~\cite{sun2020multi} adopts a node clustering task to align and refine "pseudo labels" generated by self-training. For link prediction and node clustering, Adaptive Graph Encoder (AGE)~\cite{cui2020adaptive} firstly ranks the pairwise similarity according to predicted node embedding and then sets the maximum rank of positive samples and the minimum rank of negative samples. In other words, AGE generates more “pseudo-edges” for graph embedding. 

However, self-training cannot exploit the full potential in this direction given sparse graphs: 
(1) \underline{Self-supervising quantity}. The performance of GNN  highly relies on the quantity of "pseudo-labels" and "pseudo-edges", but self-training only selects a small number of confident labels and edges during each training epoch, which hinders the ability to address the graph sparsity issue.
    (2) \underline{Self-supervising quality}. The predicted pseudo information is predicted by the trained model itself, but the performance of a single model cannot be guaranteed, especially in the sparse graph setting. 

To realize the full potential for sparse graphs, we introduce to our best knowledge, the
first dedicated online knowledge distillation approach for sparse graphs, called \sys, which distills the reception-aware teacher knowledge by proposed reception-aware graph neural networks (student models). \
Instead of pre-training a static teacher model, \sys trains simultaneously a set of student models that learn from each other in a peer-teaching manner. 
The core of \sys is the knowledge 
preserving on multi-scale reception fields, which encodes
the sparse graphs at different levels of localities and hence provides richer and complimentary supervised information
from the students with different reception-aware graph knowledge.
Meanwhile, to improve self-supervising quality in the distilled knowledge, we propose to generate a more powerful teacher by ensembling  multi-scale knowledge from students. The diversity, which is a key factor in constructing an effective ensemble teacher~\cite{Snapshot_boosting,zhang2020efficient}, is enhanced by optimizing individual reception-aware graph knowledge for each student.
For better self-supervising quantity, each student can get sufficient supervision from the reception-aware graph knowledge, task-specific knowledge, and the rich distilled knowledge (soft label) from a powerful teacher model.
We also adopt new decoupled graph neural networks as student models towards deeper graph propagation, allowing us to exploit the knowledge from more scales of reception field.

Different from traditional distillation works~\cite{chen2020online,zhang2020reliable} focusing on classification tasks, a unique feature of \sys is that it can support a variety of graph learning tasks, including node classification, link prediction,  and node clustering. Experiments on 9 datasets show that our method consistently achieves state-of-the-art performance among all the compared baselines, validating the effectiveness and generalization of ROD. We also show that our method has high efficiency as compared to strong baselines.  

In summary, the contributions of
the paper are listed below:
\begin{itemize}
    \item {\it Online distillation framework.} We propose a new online distillation framework  for sparse graphs. To the best of our knowledge, this is the first online distillation method that  tackles the edge sparsity and the label sparsity problem of GNNs, while supporting various types of graph-based tasks.
    \item {\it Reception-aware model.} We propose a new decoupled GNN which is able to learn various reception-aware graph knowledge. Compared with existing GNNs, \sys can leverage the ensemble knowledge from the multi-scale reception field.
    \item {\it Performance.} Experimental results on a wide variety of learning tasks and graph datasets demonstrate that our method outperforms all other methods consistently.
\end{itemize}

%% file: 2.pre.tex
\section{Preliminaries}
\subsection{Problem Formalization}
Throughout this paper, we consider an undirected graph $\mathcal{G}$ = ($\mathcal{V}$,$\mathcal{E}$) with $|\mathcal{V}| = N$ nodes and $|\mathcal{E}| = M$ edges. We denote by $\mathbf{A}$ the adjacency matrix of $\mathcal{G}$, weighted or not. Nodes can possibly have features vector of size $d$, stacked up in an $N \times d$ matrix $\mathbf{X}$. $\mathbf{D}=\operatorname{diag}\left(d_{1}, d_{2}, \cdots, d_{N}\right) \in \mathbb{R}^{n \times n}$ denotes the degree matrix of $\mathbf{A}$, where $d_{i}=\sum_{v_{j} \in \mathcal{V}} \mathbf{A}_{i j}$ is the degree of node $v_{i}$. In this paper, we propose a universal GNN framework that tackles both the unsupervised graph embedding and the semi-supervised node classification tasks. 

The aim of graph embedding is to map nodes to low-dimensional embeddings. We denote $\mathbf{Z}$ as the embedding matrix, and the embeddings preserve both the topological structure and feature information of graph $\mathcal{G}$. 
For downstream tasks, we consider link prediction and node clustering. The task of link prediction is to predict whether an edge exists between a pair of nodes. The node clustering task requires the model to partition the nodes into $m$ disjoint groups ${G_{1},G_{2},\cdots,G_{m}}$, where similar nodes should be in the same group. Besides, suppose $\mathcal{V}_l$ is the labeled set, the goal of semi-supervised node classification is to predict the labels for nodes in the unlabeled set $\mathcal{V}_u$ with the supervision of $\mathcal{V}_l$.

\subsection{Graph Neural Networks}
GCN ~\cite{kipf2016semi} is the most representative method in GNNs.
Based on the intuitive assumption that locally connected nodes are likely to have the same label,
GCN iteratively propagates the information of the nodes to the adjacent nodes when predicting a label.
In a GCN, each layer updates the node feature embedding in the graph by aggregating the features of neighboring nodes:

\begin{small}
\begin{equation}
    \mathbf{X}_{(l+1)}=\delta\big(\widehat{\mathbf{A}}\mathbf{X}_{(l)}\mathbf{W}^{(l)}\big),
    \label{eq_GC}
\end{equation}
\end{small}
where $\mathbf{X}^{(l)}$ and $\mathbf{X}^{(l+1)}$ are representation vectors of layer $l$ and $l+1$. $\widetilde{\mathbf{A}}=\mathbf{A}+\mathbf{I}_{N}$ is used to aggregate feature vectors of adjacent nodes, with $\mathbf{I}$ being the identity matrix.
In other words, $\widetilde{\mathbf{A}}$ is the adjacent matrix of the undirected graph $\mathcal{G}$ with added self-connections.
Note that $\widehat{\mathbf{A}}$ is the normalized adjacent matrix, and there exist lots of normalization methods\cite{klicpera2019diffusion,kipf2016semi}.
A commonly used method is the symmetric normalized Laplacian, in which $\widehat{\mathbf{A}} = \widetilde{\mathbf{D}}^{-\frac{1}{2}}\widetilde{\mathbf{A}}\widetilde{\mathbf{D}}^{-\frac{1}{2}}$.


\paragraph{\underline{Decoupled GNNs}}Recently, some studies~\cite{wu2019simplifying,he2020lightgcn} have shown that the expressive power of GNNs mainly relies on the repeated graph propagation rather than nonlinear feature extraction. Based on this observation, these methods propose to replace the repeated feature propagation with a single matrix by raising the normalized adjacency matrix to a higher power.
Then, they propagate the original node feature matrix $\mathbf{X}_0$ with the high order propagation matrix and then transform the propagated embedding with  the non-linear transformation $f_\theta$. SGC ~\cite{wu2019simplifying} is the most representative method among them, and the graph convolution layer in SGC is decoupled into nonlinear transformation and propagation. Correspondingly, the above Eq.~\eqref{eq_GC} can be decoupled into two processes:
\begin{small}
\begin{equation}
\begin{aligned}
    \mathbf{X}_k=\widehat{\mathbf{A}}^{k}\mathbf{X}_0,\ 
    \mathbf{Z}=f_\theta(\mathbf{X}^k),
    \label{aptt}
\end{aligned}
\end{equation}
\end{small}
where $\mathbf{Z}$ is the final optimized embedding, $\mathbf{X}_0$ is the original feature matrix and $f_\theta$ is used for non-linear transformation, and it is usually defined as an MLP model. Based on SGC, a series of models have attempted to decouple the propagation and transformation process while retaining the model performance.

Several recent works show that the entanglement of propagation and noon-liner transformation will harm both performance and robustness.
For example, LightGCN~\cite{he2020lightgcn} indicates this entanglement will harm the recommendation performance of Neural Graph Collaborative Filtering. With the disentangled architecture, AGE~\cite{cui2020adaptive} gets state-of-the-art performance in both link prediction and node clustering tasks. Besides, both AP-GCN ~\cite{spinelli2020adaptive} and APPNP ~\cite{klicpera2018predict} decouple the transformation and propagation, thus they can scale to very high depth and get better performance without suffering from the over smoothing problem in the node classification task.

\subsection{Reception Field of GNNs}
\begin{figure}[tpb]
    \centering
    \includegraphics[width=.8\linewidth]{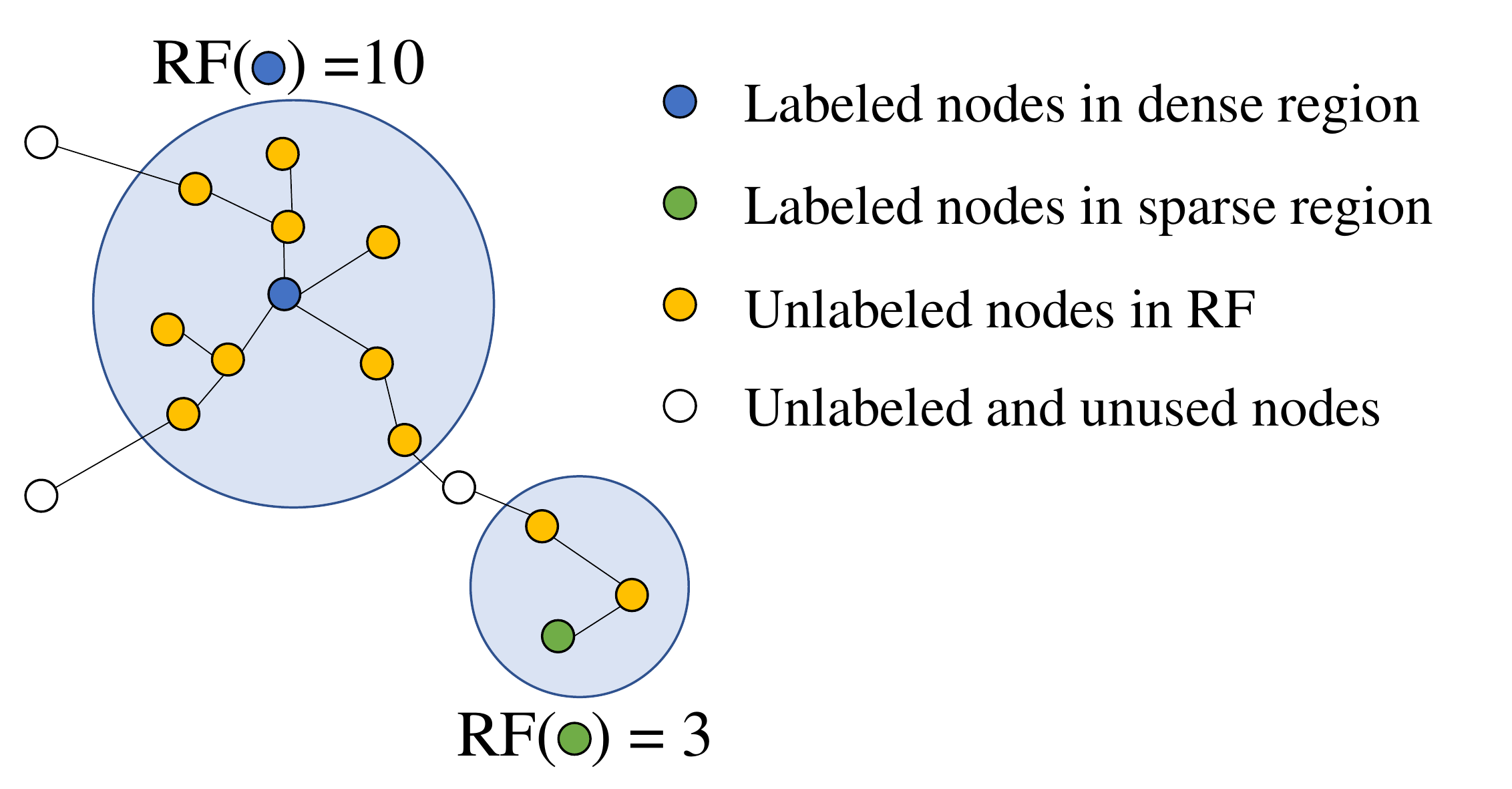}
    \vspace{-3mm}
    \caption{RF of dense and sparse nodes. }
    \vspace{-5mm}
    \label{RF}
\end{figure}

\begin{figure}[tpb]
    \centering
    \includegraphics[width=.9\linewidth]{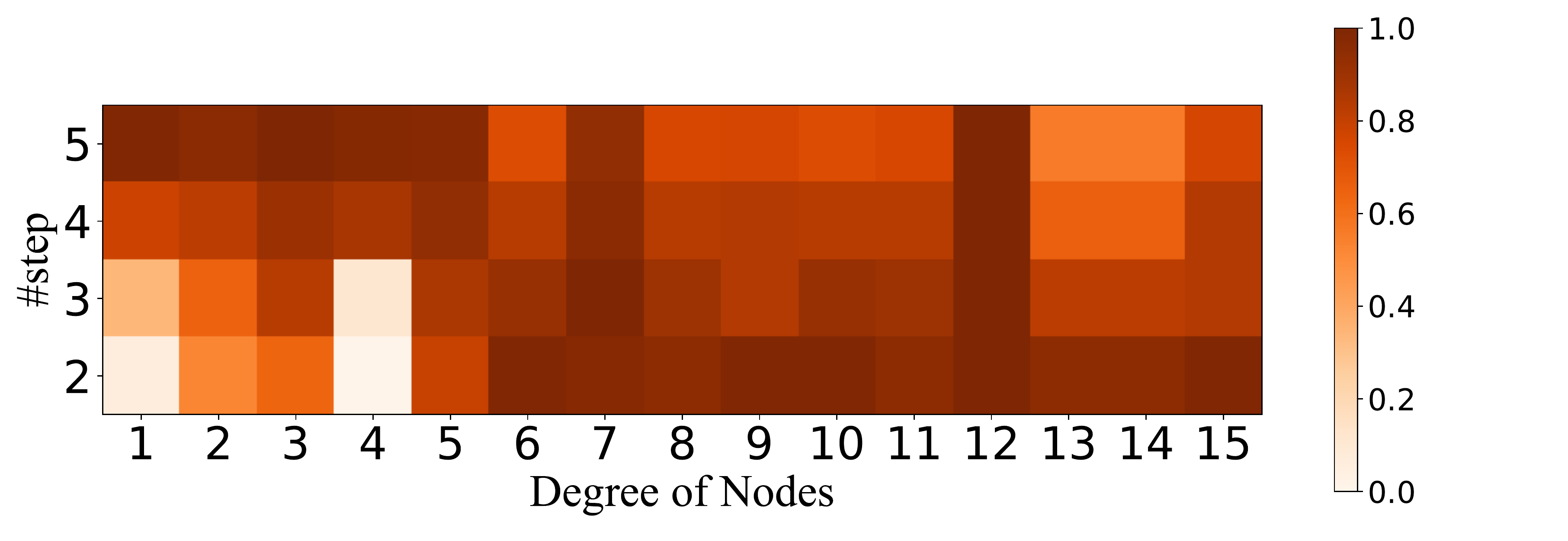}
    \vspace{-3mm}
    \caption{The effect of different propagation steps on 15 randomly selected nodes (one node for each degree) on Cora dataset.  The X-axis is the node degree and Y-axis is the propagation steps in SGC. the color represents the ratio of being correctly predicted in 100 different runs.}
    \vspace{-5mm}
    \label{degree-acc}
\end{figure}

For a $K$-layers GNN, the receptive field of a node refers to a set of nodes including the node itself and its $K$-hops neighbors (See Figure \ref{RF}). 
From the perspective of label propagation, 
as shown in Figure~\ref{RF}, labeled nodes in the sparse region get extremely small RF and it can hardly propagate its distant neighbors. Under edge sparsity and label sparsity, it is hard to sufficiently propagate the label and feature information to large RF, yielding the unsatisfying downstream task performance.  
To increase RF of low degree nodes,
a simple way is to train a deeper GNN.
However, recent studies reveal the repeated propagation in GNNs makes node representations of different classes indistinguishable, which is also called over-smoothing ~\cite{li2018deeper}.
Stacking too many graph propagation layers will highly degrade the performance of GNN, and most current state-of-art GNNs include only two layers.
We argue the performance degradation problem of deep GNN mainly is caused by the fixed-hop neighborhood and insensitive to the actual demands of nodes. This either makes that long-range dependencies cannot be fully leveraged due to limited hops/layers, or loses local information due to introducing many irrelevant nodes when increasing the number of hops.
To demonstrate this, we apply SGC with different propagation steps to conduct node classification on Cora dataset.
Figure~\ref{degree-acc} shows that the nodes with low degrees are more likely to be misclassified but they can get better performance with more propagation steps. 
As a result, a small hop may fail to capture sufficient neighborhood information for low-degree nodes, while a large hop may degrade the  performance of high-degree nodes. Interesting knowledge might exist on different levels of locality, which makes it necessary to consider multi-scale reception field.

\subsection{Self-supervised Learning}
Self-training and Online Distillation are two popular techniques in self-supervised learning~\cite{chen2020online}. 
The general self-training pipeline starts by training a model with given limited labeled data, then predicts “pseudo-labels” to all unlabeled nodes and selects the nodes with high prediction confidence, and adds these nodes to the labeled set for the next training.
In contrast to self-training which directly generates the “pseudo-labels”, Knowledge Distillation (KD) forces student models to approximate the feature representation or smooth label generated by the pre-trained teacher model. Compared with the “pseudo-labels” predicted by the student itself, the soft label predicted by the teacher model contains more information. Besides, the predicted information of the teacher is more reliable since the teacher is usually a more powerful model. Compared with self-training, Knowledge Distillation can provide supervision with higher quality.
However, such offline distillation methods rely on a strong pre-trained teacher, which requires a complex two-phase training procedure.
As a result, recent work~\cite{chen2020online,zhang2018deep} focuses on more economic online knowledge distillation, which is a specific KD method generating the distilled signal within the network itself and finishing the knowledge distillation in a one stage.

%% file: 3.method.tex
\section{The proposed method}
In this section, we introduce our proposed \sys, a novel reception-aware online distillation framework, which addresses the graph sparsity issues and improves the performance of a variety of tasks such as link prediction, node clustering, and node classification.
\begin{figure*}[htpb]
    \centering
    \includegraphics[width=0.9\linewidth]{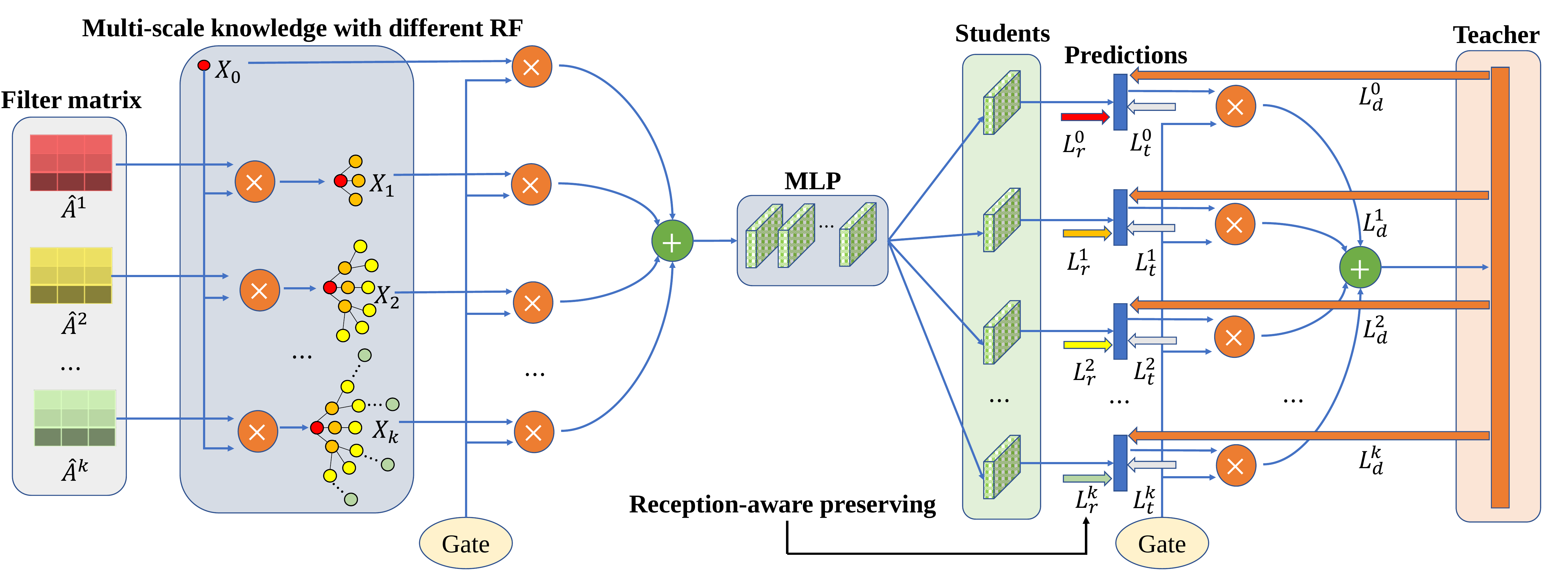}
    \vspace{-3mm}
    \caption{Overview of online distillation training of \sys.}
     \vspace{-4mm}
    \label{Fig.overview}
\end{figure*}

\subsection{Online Distillation Framework}
We illustrate the workflow of the proposed \sys online knowledge distillation framework in Figure \ref{Fig.overview}.
Given the raw feature matrix $\mathbf{X}_0$, we first precompute the aggregated feature $\mathbf{X}_k$ for the reception field within the  $k$-hop neighborhood.
Then a gate mechanism is used to adaptively combine the precomputed features aggregated over different hops, and this combined node feature will be feed into a shared MLP for feature encoder. We design three supervision signals for \sys: multi-scale reception-aware graph knowledge, task-based supervision, and rich distilled knowledge.

With the encoded multi-scale feature $\mathbf{X}_k$, we use $k$ individual layers (named "students") to preserve the information of different reception fields.
Specifically, \sys first generates the node similarity matrix based on each feature $\mathbf{X}_k$ and then enforces the student to learn a similar local
embedding similarity distribution by minimizing the distance between the similarity matrices.

Next, we get the predictions of $k$ student models and adaptively combine these predictions with a gate component to get the ensemble teacher. In this way, the teacher contains more accurate predictions than the student since it contains multi-scale reception-aware graph knowledge.
Then we distill the knowledge
back into all students in a closed-loop form. Each student will be supervised under teacher-student distillation loss. 

Finally, to improve the model performance on a specific task (i.e., node classification, clustering, and link prediction), all students will be optimized by a task-specific loss. These three loss functions for each student model are optimized in a joint learning way.

\subsection{Reception-aware knowledge Preserving}
\paragraph{\underline{Reception supervised knowledge}} We first perform $k$-step feature propagation using filter $\widehat{\mathbf{A}}^k$ to compute $\mathbf{X}_{k}=\widehat{\mathbf{A}}^k\mathbf{X}$, which denotes the representations obtained by propagating information from nodes that are
$K$-hop away.

The cosine similarity matrix $S_k$ in the $k$-th reception field is:
\begin{small}
\begin{equation}
\mathbf{S}_{k} =  \frac{\mathbf{X}_{k}\mathbf{X}_k^{T}}{{\|\mathbf{X}_{k}\|_{2}}^{2}},
\end{equation}
\end{small}
which encodes the $k$-hop neighborhood information (i.e., the $k$-th scale of the reception field). As the step $k$ increase, more
global information is included.

\vspace{-2mm}
\paragraph{\underline{Student model}} As shown in Figure \ref{Fig.overview}, each student model is an auxiliary branch, which serves as an independent model with a shared
low-level feature encoder. 
As Figure~\ref{degree-acc} shows, it is hard to find a suitable RF for each node with a fixed $k$-step propagation. The shared feature encoder uses a gate component that learns to combine node features aggregated over different levels of localities:
\begin{small}
\begin{equation}
\begin{aligned}
      & \mathbf{H}=\sum_{k=0}^K \sigma(\bm{W} \mathbf{X}_k) \cdot \mathbf{X}_k ,
\end{aligned}
\end{equation}
\end{small}
where $\mathbf{H}$ is the combined feature matrix, and $\sigma$ denotes the sigmoid function.
We then feed $\mathbf{H}$ into different student model to get the corresponding node embedding $\mathbf{Z}_k$ and node similarity $\mathbf{P}^{(k)}$ for the $k$-th student model:
\begin{small}
\begin{equation}
\mathbf{P}_{k} =  \frac{\mathbf{Z}_{k}\mathbf{Z}_k^{T}}{{\|\mathbf{Z}_{k}\|_{2}}^{2}}.
\end{equation}
\end{small}
\vspace{-2mm}
\paragraph{\underline{Reception-knowledge preserving}}
To preserve the knowledge in different scales of reception fields, we propose to reconstruct the similarity matrix $\mathcal{S}_k$ for the $k$-th student model and then define the following reception-aware information preserving loss function:
\begin{equation}
\begin{aligned}
      & \mathcal{L}_r^{(k)}= \|\mathbf{P}^{(k)}-\mathbf{S}_k\|_{F},
      \label{classi}
\end{aligned}
\end{equation}
which enables individual students to make complementary predictions based on their reception-scale knowledge.

\vspace{-2mm}
\paragraph{\underline{Reliable supervision via sampling}}
Considering the efficiency and performance issues, we propose to sample reliable positive/negative nodes based on node similarity.
Considering the high variance introduced by a single model's prediction, we propose to measure the node similarity by the ensemble of similarity matrix in different RF spaces as follows:
\begin{small}
\begin{equation}
\mathbf{S} =  \sum_{k=0}^K \mathbf{S}_{k},
\end{equation}
\end{small}
Then we rank the pairwise similarity sequence in descending order. 
We assume an edge $(v_i, v_i)$ is positive if it exists in the original graph ($\mathbf{A}_{ij} = 1$) or their similarity is as high as $\mathbf{S}_{i,j}$ is the maximum top-$M$ value in the sorted similarity sequence. A set of all these positive edges is denoted as $\mathcal{V}_{pos}$. Similarly, we select $P$ negative samples  (denoted by $\mathcal{V}_{neg}$) with the minimum top-$P$ ranks. 
Then we get a sampled adjacent matrix $\overline{\mathbf{A}}$ :
\begin{small}
\begin{equation}
\overline{\mathbf{A}}_{ij}=\left\{
\begin{aligned}
&1,       & (v_i, v_j) \in \mathcal{V}_{pos}\\
&0,       & (v_i, v_j) \in \mathcal{V}_{neg}\\
&\text{None},     & \text{otherwise}\\
\end{aligned}
\right..
\end{equation}
\end{small}

Compared with the original adjacency matrix $\mathbf{A}$, the negative samples are more reliable in $\overline{\mathbf{A}}$ since each $\mathbf{S}_k$ considers both the node features $\mathbf{X}_0$ and the graph structure $\mathbf{A}$. Besides, $\mathbf{S}$ combines the node similarity in $k$ different localities.
For large graphs, we just calculate the loss in the positive samples $\mathcal{V}_{pos}$ and negative samples $\mathcal{V}_{neg}$ to improve performance and memory usage.

\subsection{Task-based Supervision}
To get better performance with a specific task, we use different supervision signals based on different graph tasks:

\vspace{-1mm}
\paragraph{\underline{Node classification}}
We adopt the Cross-Entropy (CE) measurement between the predicted softmax outputs and the one-hot ground-truth label distributions as the objective function: 
\begin{small}
\begin{equation}
\begin{aligned}
      & \mathcal{L}_t^{(k)}= -\sum_{v_i \in \mathcal{V}_l } \mathbf{Y}_{ij} \log \mathbf{P}_{ij}^{(k)},
      \label{classi}
\end{aligned}
\end{equation}
\end{small}
where $\mathbf{Y}_i$ is a one-hot label indicator vector, and $\mathbf{P}_{i}^{(k)}$ is the $k$-th student model's predicted softmax outputs of the node $v_i$.

\vspace{-1mm}
\paragraph{\underline{Link prediction}}
Similar to GAE ~\cite{kipf2016variational}, we adopt a graph reconstruction loss that aims to minimize the difference of two nodes in the positive edges and maximize this difference in the negative samples. The difference is that we enlarge the number of positive samples ($|\mathcal{V}_{pos}| \geq M$) and significantly reduce the number of negative samples ($|\mathcal{V}_{neg}| \ll N^2 - M$ ).

As shown in Figure~\ref{Fig.overview}, the $k$-th student model can get its individual node embedding $\mathbf{Z}_k$, thus we can get the predicted nodes similarity matrix as:
\begin{small}
\begin{equation}
\mathbf{P}^{(k)} =  \frac{\mathbf{Z}_{k}\mathbf{Z}_k^{T}}{{\|\mathbf{Z}_{k}\|_{2}}^{2}},
\end{equation}
\end{small}
\vspace{-2mm}

Larger $\mathbf{S}_{ij}$ means nodes $v_i$ and $v_j$ are more similar and are more likely the positive samples. Similarly, larger $1 - \mathbf{S}_{ij}$ means $(v_i, v_j)$ have a bigger chance to be the negative samples. So, we set $\mathbf{S}_{ij}$ and $1 - \mathbf{S}_{ij}$ as the sample weights of positive and negative samples respectively, and define the following optimization function:
\begin{small}
\begin{equation}
\begin{aligned}
\mathcal{L}_t^{(k)} &= \sum_{(v_i, v_j) \in \mathcal{V}_{pos}}\! -\mathbf{S}_{ij}\overline{\mathbf{A}}_{ij} \log \mathbf{P}_{ij}^{(k)} \quad - \\
&\quad \sum_{(v_i, v_j) \in \mathcal{V}_{neg}} (1-\mathbf{S}_{ij})(1-\overline{\mathbf{A}}_{ij}) \log(1- \mathbf{P}_{ij}^{(k)}).
\end{aligned}
\end{equation}
\end{small}
\vspace{-3mm}
\paragraph{\underline{Node clustering}}
We perform K-Means ~\cite{hartigan1979algorithm} on the $k$-th student model's embedding matrix $\mathbf{Z}_k$ every $m$ ($m$ = 10 in this paper) epochs and provides the cluster centroid $\mathbf{C}_j$ for each cluster $j$. Then, the objective of this task is to make each node close to the cluster it belongs to and away from others:
\begin{small}
\begin{equation}
    \mathcal{L}_t^{(k)}= \frac{1}{N}\sum_{i=1}^N\frac{1}{q-1}\sum_{\mathbf{C}_j \ne \mathbf{Y}_i}\left\|\mathbf{Z}_{i}-\mathbf{Z}_{\mathbf{C}_j}\right\|_2 -  \frac{1}{N}\sum_{i=1}^N\left\|\mathbf{Z}_{i}-\mathbf{Z}_{\mathbf{Y}_i}\right\|_2,
\label{L_c}
\end{equation} 
\end{small}
where  $Y_i$ is the centroid of cluster node $v_i$ belongs to and $q$ is the number of classes.
With $\mathcal{L}_t^{(k)}$, each student can obtain decent embedding for clustering: nodes within the same cluster are gathered densely, and the boundaries between different clusters are distinct.

\subsection{Reception-aware Online Distillation}
We use an online distillation method to tackle the self-supervising quality problem (see section 2.4) of existing self-training methods. As each student preserves knowledge of different scales of reception fields, we use an ensemble method to construct the powerful teacher model for distillation. 
The ensemble teacher model preserves multi-scale reception-aware signals, which are voted by student models from different perspectives of locality levels. Besides, diversity, which is a key factor in building a strong ensemble model, is also enhanced due to different supervision signals.

Specifically, we combine the predictions of each student model and get the ensemble teacher model as:
\begin{small}
\begin{equation}
\begin{aligned}
      & \mathbf{P}^{e}=\sum_{k=0}^K \sigma(\bm{W} \mathbf{P}^{k}) \cdot \mathbf{P}^{k}, 
\end{aligned}
\end{equation}
\end{small}
where we also use a gate component that learns to ensemble all
K student branches to build a stronger teacher model.

Here, $\mathbf{P}$ is the softmax outputs ($\mathbf{P} \in \mathbb{R}^{N \times q}$) for node classification. For node clustering task $\mathbf{P}$ is defined as:
\begin{small}
\begin{equation}
    \mathbf{P}_{ij}=\frac{exp(-\mathbf{D}_{ij})}{\sum_{j=1}^{q}exp(-\mathbf{D}_{ij})},
\end{equation}
\end{small}
where $\mathbf{D}_{ij}$ is the euclidean distance between the representation of the $i$th node and the $j$th cluster centroid, $q$ is the number of clusters.
As for link prediction, $\mathbf{P}$ is denoted as the predicted node similarity matrix ($\mathbf{P} \in \mathbb{R}^{N \times N}$).

Then we minimize the Kullback-Leibler (KL) divergence between the $k$-th student model and teacher model as:
\begin{small}
\begin{equation}
\begin{aligned}
      & \mathcal{L}_d^{(k)} = KL(\mathbf{P}^{e}, \mathbf{P}^{k}) = \sum_{(i,j)}\mathbf{P}_{ij}^{e} \log \frac{\mathbf{P}_{ij}^{e}}{\mathbf{P}_{ij}^{k}}.
      \label{classi}
\end{aligned}
\end{equation}
\end{small}

\subsection{Joint Objective Function}
Finally, we jointly optimize the task-specific loss $\mathcal{L}_t$, the reception aware information preserving loss $\mathcal{L}_r$, and the distillation loss $\mathcal{L}_d$. 
The total objective function is defined as:
\begin{small}
\begin{equation}
    L=\sum_{k=0}^K \mathcal{L}_t^{(k)} + \alpha\mathcal{L}_r^{(k)} + \beta\mathcal{L}_d^{(k)},
\end{equation}
\end{small}
where coefficients $\alpha>0$ and $\beta>0$ are the balancing factors for the multiple tasks. To optimize the weighted sum of their losses,
all students share the same feature encoder (MLP in Figure~\ref{Fig.overview}) but have their individual parameters for optimizing the given task with different locality level features.

%% file: 4.experiment.tex
\section{experiment}
In this section, we empirically evaluate our framework and present complete results here.

\begin{table*}[t]
\small
\centering
\caption{Overview of the Datasets} \label{Dataset}
\vspace{-3mm}
\begin{tabular}{ccccccccc}
\toprule
\textbf{Dataset}&\textbf{\#Nodes}& \textbf{\#Features}&\textbf{\#Edges}&\textbf{\#Classes}&\textbf{\#Train/Val/Test}&\textbf{Task type}&\textbf{Description}\\
\midrule
Cora& 2,708 & 1,433 &5,429&7& 140/500/1000 & Transductive&citation network\\
Citeseer& 3,327 & 3,703&4,732&6& 120/500/1000 & Transductive&citation network\\
Pubmed& 19,717 & 500 &44,338&3& 60/500/1000 & Transductive&citation network\\
\midrule
Amazon Computer& 13,381  & 767& 245,778 & 10 &200/300/12881&Transductive&co-purchase graph\\
Amazon Photo &7,487  & 745& 119,043 & 8 & 160/240/7,087&Transductive&co-purchase graph\\
Coauthor CS& 18,333  & 6,805 & 81,894 & 15& 300/450/17,583&Transductive&co-authorship graph \\
Coauthor Physics& 34,493 & 8,415 & 247,962 & 5& 100/150/34,243&Transductive&co-authorship graph \\
\midrule
Flickr& 89,250 & 500 & 899,756 & 7 &  44,625/22,312/22,312 & Inductive &image network\\
Reddit& 232,965 & 602 & 11,606,919 & 41 &  155,310/23,297/54,358 & Inductive&social network \\
\bottomrule
\label{transductive}
\end{tabular}
\vspace{-4mm}
\end{table*}

\subsection{Datasets \& Baselines}
\paragraph{Datasets}
We conduct the experiments on public partitioned datasets, including three citation networks (Citeseer, Cora, and PubMed) in ~\cite{kipf2016semi},  four co-authorship graphs (Amazon and Coauthor) in ~\cite{shchur2018pitfalls}, and two social networks (Flickr and Reddit) in ~\cite{zeng2019graphsaint}.
Table ~\ref{Dataset} provides the overview of the 9 datasets and the detailed description is in Section A.1 of the supplemental material.
\paragraph{Baselines}
We compare ROD with these methods:
\begin{itemize}
    \item \textbf{ Node classification:} GCN ~\cite{kipf2016semi}, GAT ~\cite{velivckovic2017graph}, simplified GCN (SGC) ~\cite{wu2019simplifying}, GraphSAGE ~\cite{hamilton2017inductive}, JK-Net ~\cite{xu2018representation}, APPNP ~\cite{klicpera2018predict},   AP-GCN ~\cite{spinelli2020adaptive},FastGCN ~\cite{chen2018fastgcn}, Scalable Inception Graph Neural Networks (SIGN) ~\cite{rossi2020sign}, ClusterGCN ~\cite{chiang2019cluster} and GraphSAINT ~\cite{zeng2019graphsaint}.
    \item \textbf{ Link prediction:} 
    GAE and VGAE ~\cite{kipf2016variational}, LightGCN ~\cite{he2020lightgcn}, simplified GCN (SGC) ~\cite{wu2019simplifying}, Scalable Inception Graph Neural Networks (SIGN) ~\cite{rossi2020sign}, Adaptive Graph Encoder (AGE) ~\cite{cui2020adaptive}, Spectral Clustering (SC) ~\cite{ng2001spectral} and DeepWalk (DW) ~\cite{perozzi2014deepwalk}.
    \item \textbf{ Node clustering:} GAE and VGAE ~\cite{kipf2016variational}, MGAE ~\cite{wang2017mgae}, ARGA and ARVGA ~\cite{pan2018adversarially},  AGC ~\cite{zhang2019attributed}, DAEGC ~\cite{wang2019attributed}, SDCN ~\cite{bo2020structural} and AGE ~\cite{cui2020adaptive}.
\end{itemize}
A detailed introduction of these baseline methods can be found in Section ~\ref{baseline_methods} of the Appendix.  

\subsection{Evaluation Metrics \& Parameter Settings}
For node classification, we consider the accuracy of the test data.  
For link prediction, we partition the graph following GAE and AGE, and report Area Under the Receiver Operating Characteristics (ROC) Curve (AUC) and Average Precision (AP) scores.
For node clustering, we compare three metrics: Accuracy (ACC), Normalized Mutual Information (NMI), and Adjusted Rand Index (ARI).
For ROD and other baselines, we use grid-search or follow the original papers to get the optimal hyperparameters.
More details can be found in Section A.3 of the supplementary material.

\begin{table}[t]
\caption{Test accuracy (\%) in transductive settings. Models in the first column are the coupled GNN, and the five GNN models in second column are decoupled.} 
\vspace{-3mm}
\centering
{
\noindent
\renewcommand{\multirowsetup}{\centering}
\resizebox{0.65\linewidth}{!}{
\begin{tabular}{c|ccc}
\toprule\textbf{Models}&\textbf{Cora}& \textbf{Citeseer}&\textbf{PubMed}\\
\midrule
GCN& 81.8$\pm$0.5 & 70.8$\pm$0.5 &79.3$\pm$0.7 \\
GAT& 83.0$\pm$0.7 & \underline{72.5$\pm$0.7} &79.0$\pm$0.3 \\
JK-Net& 81.8$\pm$0.5  & 70.7$\pm$0.7 & 78.8$\pm$0.7  \\
\midrule
APPNP& 83.3$\pm$0.5 & 71.8$\pm$0.5 & \underline{80.1$\pm$0.2}\\
AP-GCN& \underline{83.4$\pm$0.3}& 71.3$\pm$0.5& 79.7$\pm$0.3\\
SGC & 81.0$\pm$0.2 & 71.3$\pm$0.5 & 78.9$\pm$0.5\\
SIGN& 82.1$\pm$0.3 & 72.4$\pm$0.8 &79.5$\pm$0.5 \\
ROD& \textbf{84.7$\pm$0.6} & \textbf{73.4$\pm$0.5} &\textbf{81.1$\pm$0.4}\\ 
\bottomrule
\end{tabular}}}
\vspace{-2mm}
\label{Transductive1}
\end{table}

\begin{table}[t]
\caption{Test accuracy (\%) in transductive settings.} 
\vspace{-3mm}
\centering
{
\noindent
\renewcommand{\multirowsetup}{\centering}
\resizebox{0.8\linewidth}{!}{
\begin{tabular}{c|cccc}
\toprule\textbf{Models}&{\textbf{\makecell{Amazon \\Computer}}} & 
{\textbf{\makecell{Amazon \\Photo}}} & 
{\textbf{\makecell{Coauthor \\CS}}}&
{\textbf{\makecell{Coauthor \\Physics}}}\\
\midrule
GCN& 82.4$\pm$0.4 & 91.2$\pm$0.6 & 90.7$\pm$0.2 & 92.7$\pm$1.1 \\
GAT& 80.1$\pm$0.6 & 90.8$\pm$1.0 & 87.4$\pm$0.2 & 90.2$\pm$1.4 \\
JK-Net&82.0$\pm$0.6& 91.9$\pm$0.7 & 89.5$\pm$0.6 & 92.5$\pm$0.4 \\
\midrule
APPNP& 81.7$\pm$0.3&91.4$\pm$0.3&\underline{92.1$\pm$0.4}&92.8$\pm$0.9\\
AP-GCN& \underline{83.7$\pm$0.6}& \underline{92.1$\pm$0.3}& 91.6$\pm$0.7& \underline{93.1$\pm$0.9}\\
SGC & 82.2$\pm$0.9&91.6$\pm$0.7&90.3$\pm$0.5& 91.7$\pm$1.1\\
SIGN& 83.1$\pm$0.8&91.7$\pm$0.7&91.9$\pm$0.3& 92.8$\pm$0.8 \\
ROD& \textbf{85.3$\pm$0.8}&\textbf{93.2$\pm$0.7}  &\textbf{93.4$\pm$0.5}& \textbf{94.1$\pm$0.7}\\ 
\bottomrule
\end{tabular}}}
\vspace{-2mm}
\label{Transductive2}
\end{table}

\begin{table}[t]
\caption{ Test accuracy (\%) in inductive settings.}
\vspace{-3mm}
\centering
{
\noindent
\renewcommand{\multirowsetup}{\centering}
\resizebox{0.6\linewidth}{!}{
\begin{tabular}{c|cc}
\toprule
\textbf{Models}& \textbf{Flickr}&\textbf{Reddit}\\
\midrule
GraphSAGE & 50.1$\pm$1.3 & 95.4$\pm$0.0 \\
FastGCN & \underline{50.4$\pm$0.1} & 93.7$\pm$0.0 \\
ClusterGCN & 48.1$\pm$0.5 & 95.7$\pm$0.0 \\
GraphSAINT & 51.1$\pm$0.1 & \underline{96.6$\pm$0.1}\\
\midrule
ROD& \textbf{53.4$\pm$0.5}  & \textbf{96.7$\pm$0.1}  \\
\bottomrule
\end{tabular}}}
\vspace{-4mm}
\label{Inductive}
\end{table}

\subsection{End-to-End Comparison} 

\paragraph{\underline{Node classification}}
The classification results on three citation networks are shown in Table~\ref{Transductive1}. We observe that ROD has achieved significant improvement over all other baseline methods. Notably, ROD exceeds the current state-of-the-art model APPNP by a large margin of 1.0\% on the largest citation networks PubMed.
We further evaluate ROD on the commonly used co-authorship and co-purchase datasets, and the results are summarized in Table~\ref{Transductive2}. 
ROD outperforms the best baseline by significant margins of 1.6\%, 1.1\%, 1.3\%, and 1.0\% on the Amazon Computers, Amazon Photo,  Coauthor CS, and Coauthor Physics, respectively. 

Compared with the GCN and its variants in the first column of Table~\ref{Transductive1} and Table~\ref{Transductive2}, the decoupled methods (i.e., APPNP, AP-GCN) get better performance, which is consistent with AGE's argument that the entanglement of representation transformation and propagation would compromise the performance. 
Compared with these strong decoupled baselines, ROD still shows extremely better performance. Both APPNP and AP-GCN are able to adaptively alter reception fields either by using an additional union to decide the extent of reception or directly superpose the layer outputs of different hops.
The significant improvement of ROD mainly relies on our reception-aware information preserving and online distillation. 

To measure the generalization ability of ROD, we compare it with current state-of-the-art GNN methods in inductive settings. The experiment results in Table~\ref{Inductive} shows that ROD consistently outperforms the baselines. It outperforms the best baselines FastGCN by a large margin of 3.0\% on the widely-used Flickr dataset, demonstrating that ROD is powerful in predicting unseen nodes.

\paragraph{\underline{Link prediction}}
\begin{table}[tbp]
\centering
{
\noindent
\caption{Link prediction task in citation networks}
\vspace{-3mm}
\label{link_pre_performance}
\renewcommand{\multirowsetup}{\centering}
\resizebox{0.8\linewidth}{!}{
\begin{tabular}{c|cc|cc|cc}
\toprule
\multirow{2}{*}{\textbf{Models}} & \multicolumn{2}{c|}{\textbf{Cora}} & \multicolumn{2}{c|}{\textbf{Citeseer}} & \multicolumn{2}{c}{\textbf{PubMed}} \\ \cline{2-7} 
                         & \textbf{AUC}    & \textbf{AP}    & \textbf{AUC}    & \textbf{AP}      & \textbf{AUC}     & \textbf{AP}\\ 
\midrule
SC& $84.6$ & $88.5$ & $80.5 $ & $85.0 $ & $84.2 $ & $87.8 $ \\
DW& 83.1 & 85.0  & 80.5& 83.6 & 84.4 & 84.1  \\
GAE&91.0& 92.0& 89.5& 89.9& \textbf{96.4} & \textbf{96.5} \\
VGAE & $91.4$ & $92.6$ & $90.8 $ & $92.0 $ & $94 . 4$ & $9 4 . 7 $ \\
LightGCN & $88.4$ & $89.2 $ & $85.9 $ & $87.1$ & $93.1 $ & $93.6$ \\
SGC &87.3 &88.6 &85.4&86.1 &92.4 &93.5\\
SIGN &88.6 &90.1 &86.3&87.7 &93.5 &94.5 \\
AGE &\underline{95.7} &\underline{95.2}&\underline{96.4}&\underline{96.8} &$93.9$ &$93.1$ \\
\midrule
ROD &$\textbf{96.0} $&$\textbf{96.4}$&$\textbf{97.1}$&\textbf{97.3} &\underline{94.9} &\underline{95.1} \\
\bottomrule
\end{tabular}}}
\vspace{-4mm}
\end{table}

\begin{figure*}
 \centering
 \subfigure[ACC]{
  \scalebox{0.32}[0.32]{
   \includegraphics[width=1\linewidth]{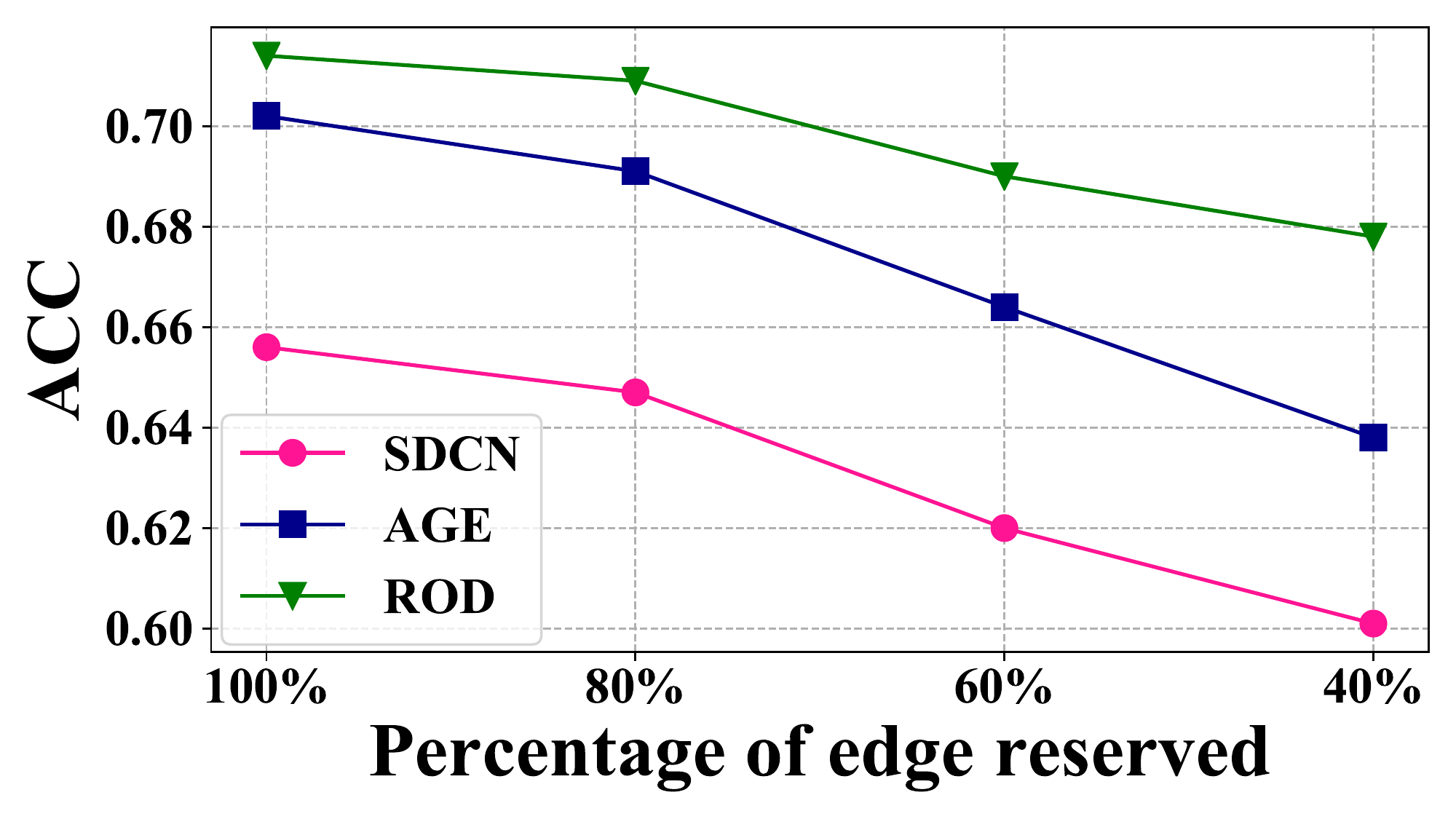}
 }}
 \subfigure[NMI]{
  \scalebox{0.32}[0.32]{
   \includegraphics[width=1\linewidth]{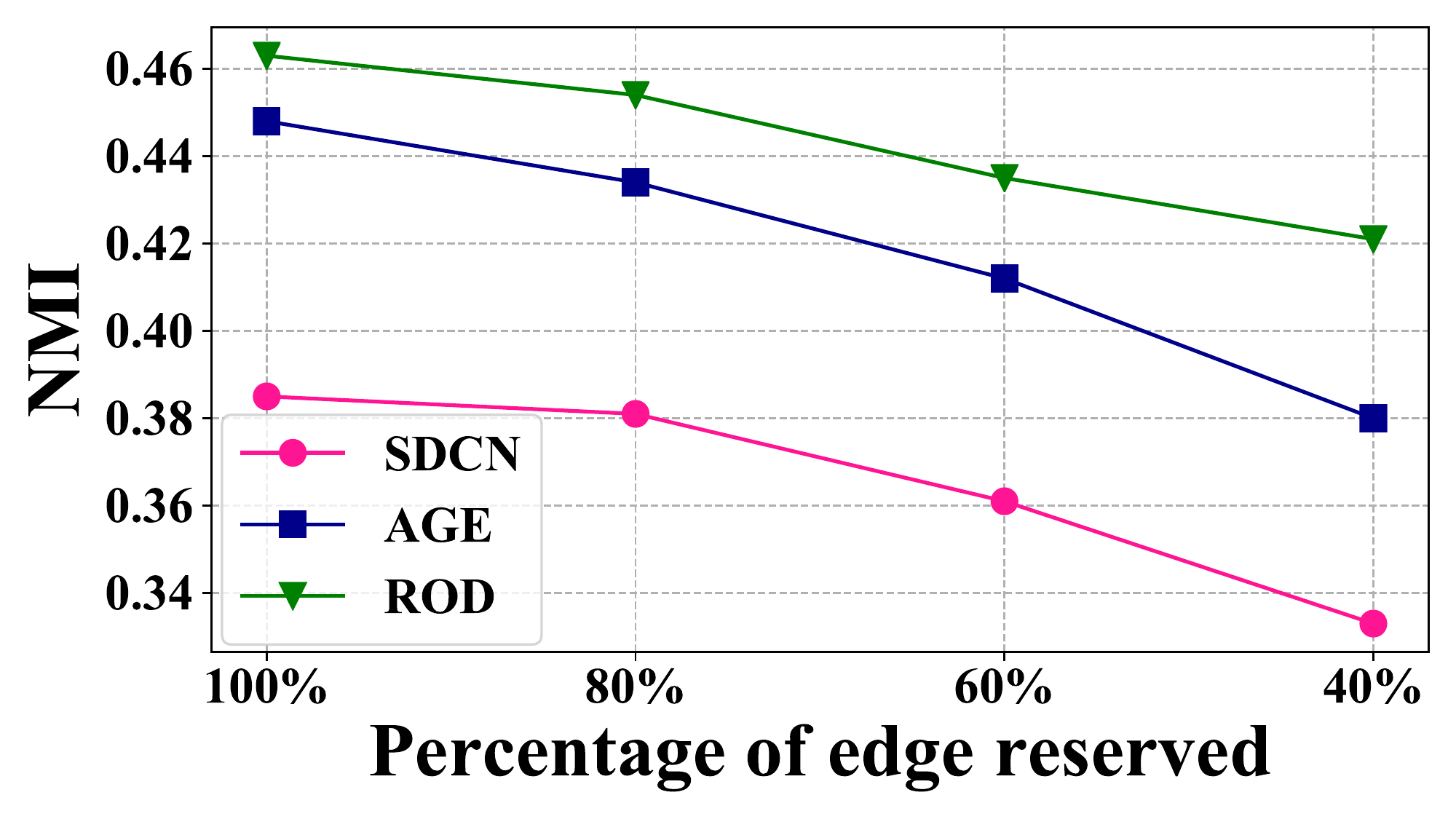}
 }}
 \subfigure[ARI]{
  \scalebox{0.32}[0.32]{
   \includegraphics[width=1\linewidth]{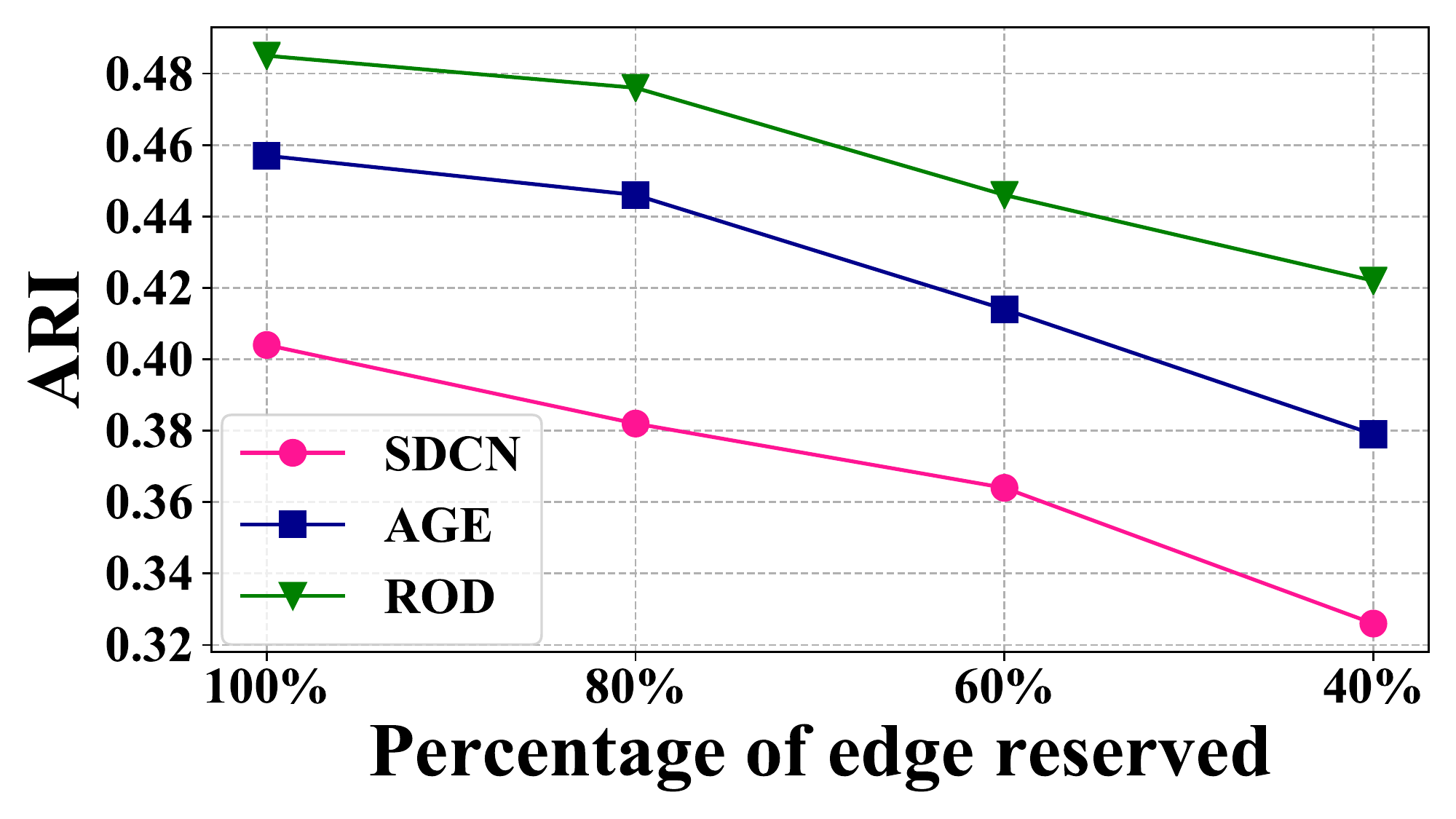}
 }}
 \vspace{-3mm}
\caption{The clustering performance along with different percentage of reserved edge on Citeseer dataset.}
\vspace{-2mm}
\label{fig:unscale}
\end{figure*}

Table ~\ref{link_pre_performance} further shows the performance of different methods on the link prediction task. Apparently, ROD performs the best baseline in two datasets among three citation networks. 
Compared with the state-of-the-art method AGE, though both ROD and AGE combine information from feature and graph structure and generate the pseudo edge information in a self-supervised way, our ROD outperforms AGE in terms of AUC by 0.3\%-1.0\% and AP by 0.5\%-2.0\%. 
The improvement can be explained by the enforcement in self-supervising quantity and self-supervising quality.
For the self-supervising quality, ROD gets extra supervision from the multi-scale reception-aware graph knowledge and a powerful teacher model. For the self-supervising quantity, the pseudo edge information of ROD is generated by a powerful teacher model, and the information is used in soft label format. In contrast, AGE only predicts hard labels by itself and such hard label loses the difference information within positive or negative edges.

Compared with decoupled GNN methods like SGC, LightGCN, and SIGN, ROD also shows a better performance. Specifically, SIGN uses different operations to gather information, which has similar insight with ROD in collecting multi-aspect information. However, the AUC of ROD exceeds SIGN by a large margin of 10.8\% on the Citeseer dataset, due to ROD's strong representation ability from the reception-aware information preserving and online distillation. 
\begin{table}[tbp]
\centering
{
\noindent
\caption{\small Node clustering task in citation networks.}
\vspace{-3mm}
\label{clustering_results}
\renewcommand{\multirowsetup}{\centering}
\resizebox{\linewidth}{!}{
\begin{tabular}{c|ccc|ccc|ccc}
\toprule
\multirow{2}{*}{\textbf{Models}} & \multicolumn{3}{c|}{\textbf{Cora}} & \multicolumn{3}{c|}{\textbf{Citeseer}} & \multicolumn{3}{c}{\textbf{PubMed}} \\ \cline{2-10} 
                         & \textbf{ACC}    & \textbf{NMI}    & \textbf{ARI}    & \textbf{ACC}      & \textbf{NMI}     & \textbf{ARI}     & \textbf{ACC}     & \textbf{NMI}     & \textbf{ARI}    \\ 
\midrule
GAE                      & 53.3  & 40.7  & 30.5  & 41.7    & 18.1   & 12.9   & 63.1   & 24.9   & 21.7  \\
VGAE                     & 56.0  & 38.5  & 34.7  & 44.4    & 22.7   & 20.6   & 65.5   & 25.0   & 20.3  \\
MGAE                     & 63.4  & 45.6  & 43.6  & 63.5    & 39.7   & 42.5   & 59.3   & 28.2   & 24.8  \\
ARGA                     & 63.9  & 45.1  & 35.1  & 57.3    & 35.2   & 34.0   & 68.0   & 27.6   & 29.0  \\
ARVGA                    & 64.0  & 44.9  & 37.4  & 54.4    & 25.9   & 24.5   & 51.3   & 11.7   & 7.8  \\
AGC                      & 68.9  & 53.7  & 48.6  & 66.9    & 41.1   & 41.9   & 69.8   & \underline{31.6}   & \underline{31.8}  \\
DAEGC                    & 70.2  & 52.6  & 49.7  & 67.2    & 39.7   & 41.1   & 66.8   & 26.6   & 27.7  \\
SDCN                     & 67.2  & 51.3  & 46.4  & 65.6    & 38.5   & 40.4   & 66.2   & 25.3   & 26.8\\
AGE                      & \underline{72.8}  & \underline{58.1}  & \textbf{56.3}  & \underline{70.2}    & \underline{44.8}   & \underline{45.7}   & \underline{69.9}   & 30.1   & 31.4  \\
\midrule
ROD  & \textbf{74.9}        & \textbf{58.3}           & \underline{54.0}       & \textbf{71.4}        & \textbf{46.3}       & \textbf{48.5}          & \textbf{70.2}   & \textbf{31.7}   & \textbf{32.2} \\
\bottomrule
\end{tabular}}}
\vspace{-4mm}
\end{table}

\vspace{-1mm}
\paragraph{\underline{Node Clustering}}
\label{node_clustering_perf}
The node clustering results of each method are shown in Table ~\ref{clustering_results}.
ROD consistently outperforms the strongest baseline AGE on the three citation datasets. 
For example, the accuracy of ROD exceeds AGE by 2.1\%, 1.2\%, and 0.3\% in Cora, Citeseer, and PubMed respectively.
Similar to the reason in the above link prediction task, we attribute this improvement to the self-supervising quality and self-supervising quantity.
Besides, ROD significantly outperforms the unsupervised learning methods GAE and VGAE on three citation networks, which proves the effectiveness of decoupled GNN on node representation learning.

\subsection{Influence on Edge Sparsity}
\begin{figure*}
 \centering
 \subfigure[Edge sparsity]{
 \label{edge_sparsity}
  \scalebox{0.32}[0.32]{
   \includegraphics[width=1\linewidth]{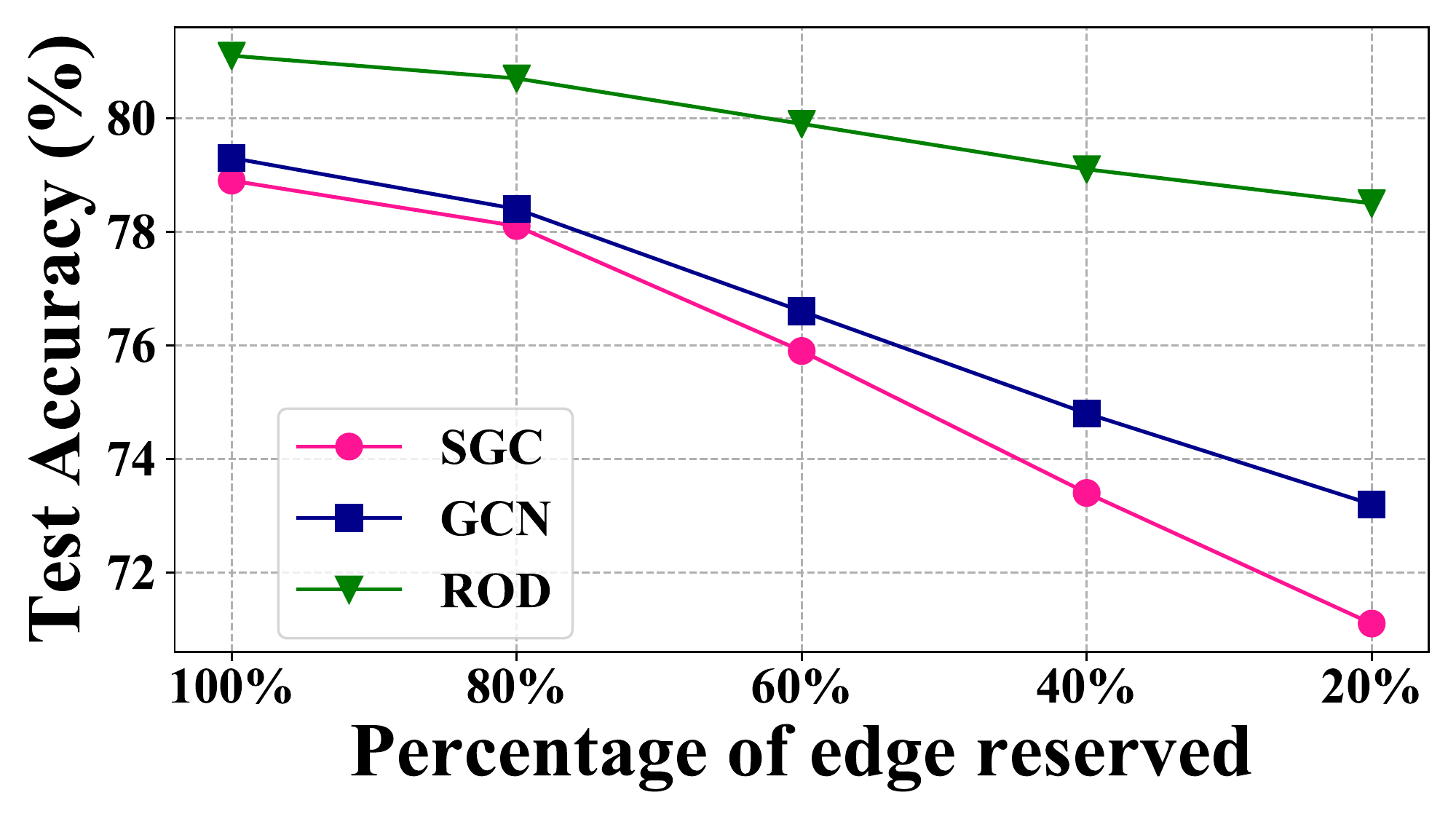}
 }}
 \subfigure[Label sparsity]{
 \label{label_sparsity}
  \scalebox{0.32}[0.32]{
   \includegraphics[width=1\linewidth]{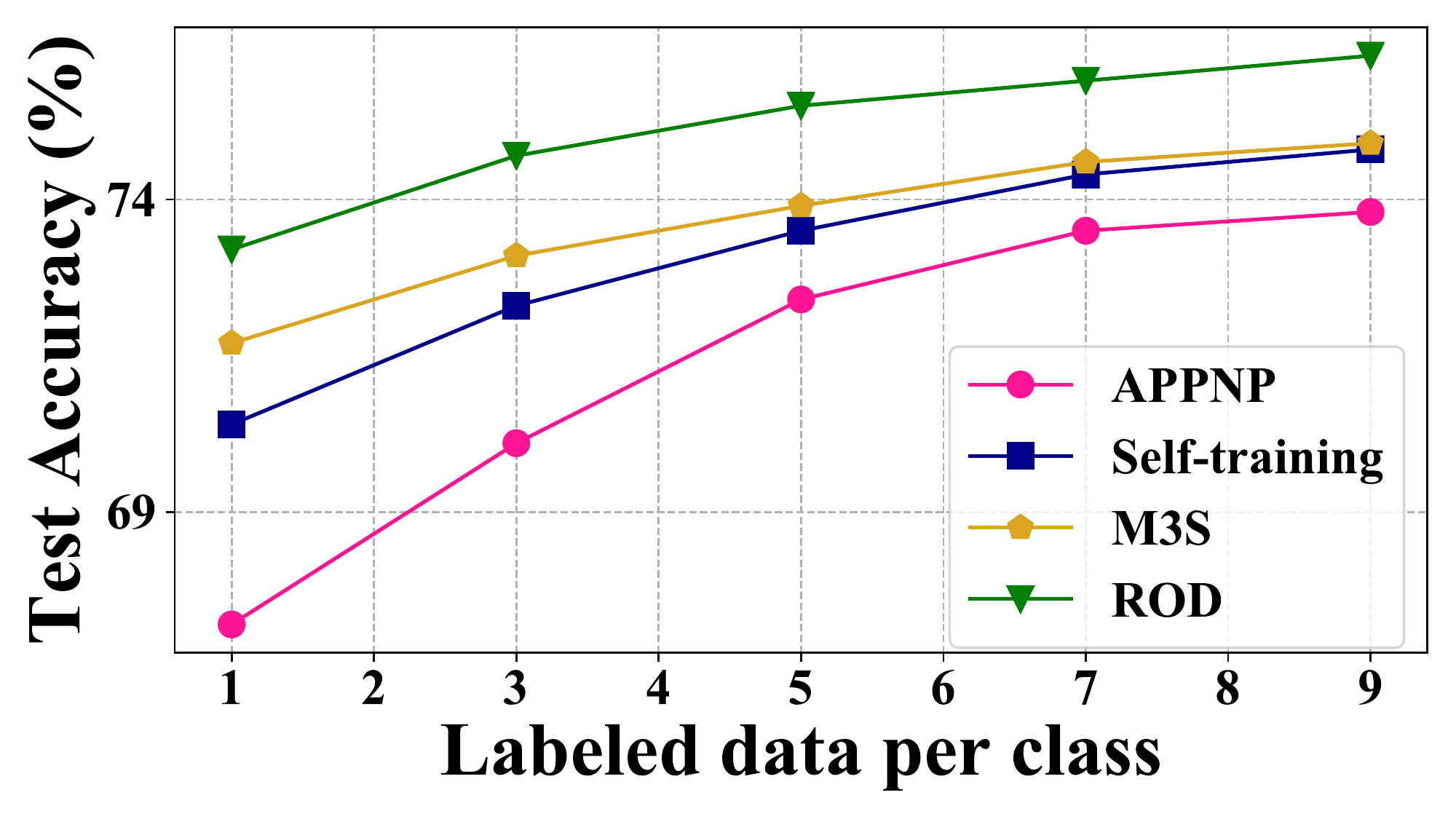}
 }}
 \subfigure[Propagation depth]{
 \label{prop_depth}
  \scalebox{0.32}[0.32]{
   \includegraphics[width=1\linewidth]{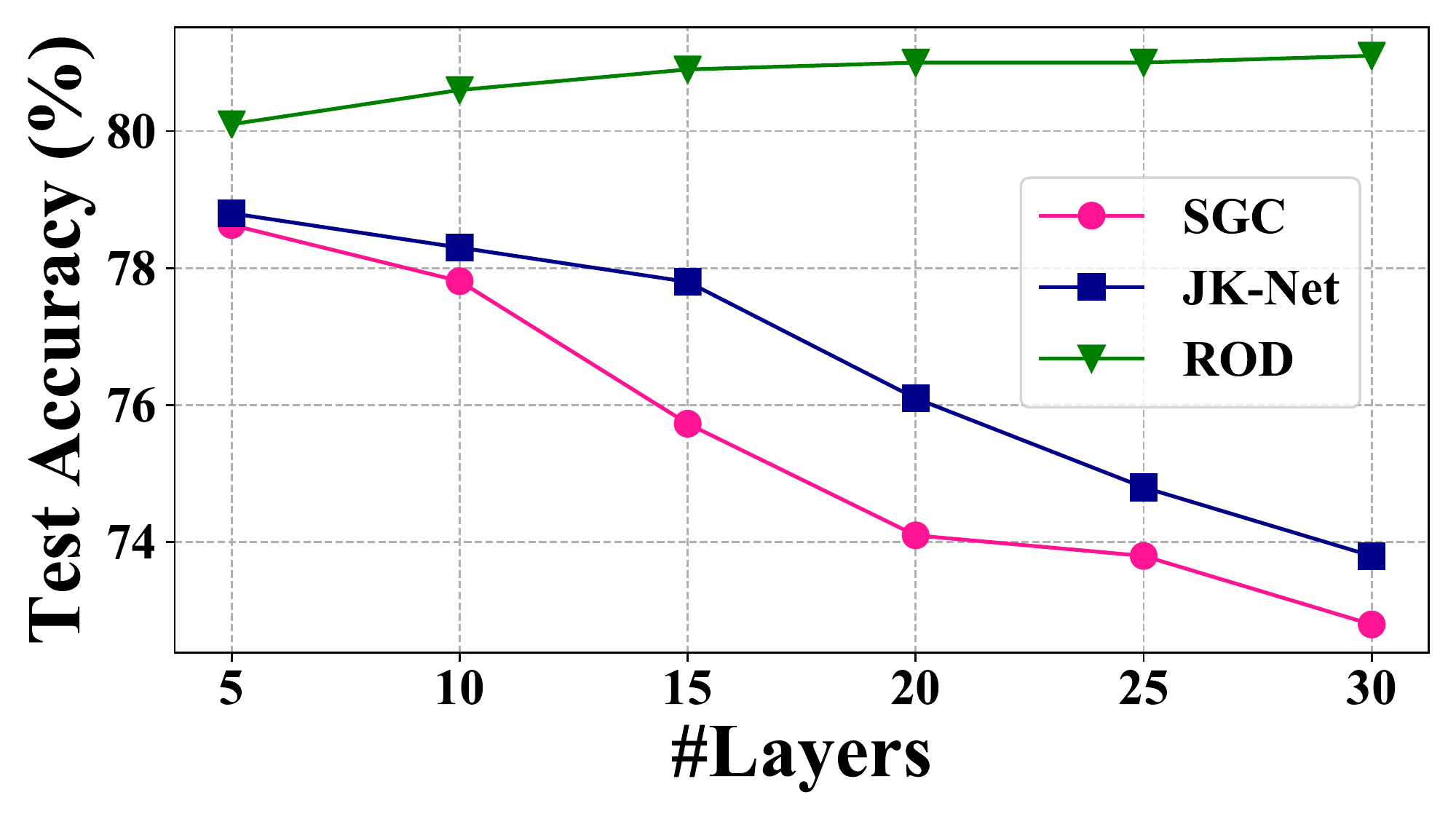}
 }}
 \vspace{-5mm}
\caption{The test accuracy along with different edge sparsity, label sparsity and propagation depth on PubMed dataset.}
\vspace{-4mm}
\label{fig:classi}
\end{figure*}

A key advantage of ROD is its ability in handling edge sparsity. It is interesting to investigate how ROD performs on different sparsity.
To do so, we randomly remove a fixed percentage of edges from the original graph, simulating various edge sparsity settings.
We compare the performance of ROD with strong baseline methods on the node clustering and node classification tasks.
For node clustering, we compare ROD with AGE and SDCN on Citeseer dataset with different percentages of reserved edges.
The variation trend of the three node clustering evaluation metrics ACC, NMI, and ARI are reported in Figure~\ref{fig:unscale}.
For node classification, we compare ROD with GCN and SGC on PubMed dataset, and the corresponding test accuracy is provided in Figure~\ref{edge_sparsity}.

As shown in Figure~\ref{fig:unscale}, despite that the performance of all the three methods degrades as the percentage of reserved edges drops, ROD has the smallest degraded margin of performance, compared with both AGE and SDCN.
For example, when 40\% of the edges are reserved, the dropped margins of clustering accuracy for AGE and SDCN are 9.1\% and 8.4\%, respectively, whereas the dropped margin of ROD is only 5.0\%.
Besides, both GAE and ROD consistently outperform SDCN in these three evaluation metrics. 
This is because these two methods can generate extra self-supervised pseudo edge information. 
Especially, due to the high quality and large quantity of the pseudo data, ROD outperforms AGE in most circumstances (except for ARI in Cora dataset).  

Only with a sufficient number of edges can the model propagates its small amount of label signal to a larger region of the entire graph and incorporate more unlabeled nodes in the model training. 
Figure~\ref{edge_sparsity} indicates ROD consistently outperforms the two baselines SGC and GCN, and the performance gain is increased when the percentage of edge reserved gradually decreases. This is because ROD can get the self-supervised pseudo edge information with high quality through the reception-aware information preserving and online distillation, and ROD benefits more from the pseudo information under higher edge sparsity.
To sum up, the results of these two experiments clearly show that ROD is more robust against the edge sparsity problem than the strong baselines.

\subsection{Influence on Label Sparsity}
We conduct experiments to examine the ability of ROD to alleviate the label sparsity problem under a semi-supervised node classification task.
The public split of PubMed dataset chooses 20 nodes per class as the training set.
To tackle the label sparsity problem, we vary the nodes per class in the training set from 1 to 9, and the test set accuracy of ROD is compared with strong baselines APPNP, Self-training, and M3S.
Both the Self-training model are two representative methods for generating the self-supervised signal, and they are based on the strongest baseline APPNP.
The variation trend of the test set accuracy of ROD and the other three baseline methods are shown in Figure~\ref{label_sparsity}.

From  the figure we see that ROD outperforms all the baseline methods by a considerable margin across all label settings. 
Besides, the performance gain between APPNP and ROD is larger in the few label setting, since the labeled nodes are hard to expand their RF rapidly win APPNP.
The two self-supervised learning strategies, Self-training and M3S alleviate the label sparsity problem in APPNP. However, \sys still outperforms these methods by a significant margin. Because these methods can only get the extra supervision signals from the hard pseudo label predicted by a single model itself. By contrast, each student model in ROD can benefit from both soft label and reception-aware knowledge preserved with high quality.
This experiment demonstrates the great robustness of our model against the label sparsity problem.

\subsection{Influence on Model Depth} 
A large propagation steps/layers could leverage a larger neighborhood to deal with the graph sparsity issue. To measure how the model depth influences ROD, we compare ROD with SGC and JK-Net allowing deep propagation steps/layers, and report their test accuracy along with different model layers on the PubMed dataset. 

The results in Figure ~\ref{prop_depth} show that depth brings stable gain in the performance of ROD, but deteriorates performance of both coupled method JK-Net and decoupled method SGC. This is because each node in ROD can combine and leverage multi-scale RF in a node adaptive manner.
The performance of SGC is worse than JK-Net as the network goes deep. The reason behind this is that for a given depth $k$, SGC only captures information in the certain $k$-th dense layer, while JK-Net takes advantage of a jumping strategy, combing information range from different layers and contains more multi-scale information. 
When the depth of $k$ increases, ROD becomes more powerful as nodes in the sparse region can enhance its representation while nodes in the dense region can adaptively filter the noise generated by over-smoothing.
Although JK-Net captures the multi-scale RF knowledge, it just concatenates the knowledge from different RF and each node still cannot find its suitable RF. As a result, increasing the layers of JK-Net still cause a severe over-smoothing problem, leading to the performance degradation.

\subsection{Ablation Study}
To thoroughly evaluate our method, we provide here ablation and performance studies on link prediction and node clustering, to analyze the effectiveness of each component in ROD.

We test ROD without: {\it (i)} task-based supervision ("ROD-no-$\mathcal{L}_t$")
{\it (ii)} the reception aware information preservation loss ("ROD-no-$\mathcal{L}_r$"). Specifically, the difference between the  original high order filter similarity matrix and node similarity matrix.
{\it (iii)} the distillation loss which measures the difference between the outputs generated by the teacher model and student model ("ROD-no-$\mathcal{L}_d$").
The test results on Cora dataset are shown in Table ~\ref{ablation_cora}.

\begin{table}[tbp]
\centering
{
\noindent
\caption{Ablation study on Cora.}
\vspace{-3mm}
\label{ablation_cora}
\renewcommand{\multirowsetup}{\centering}
\resizebox{0.95\linewidth}{!}{
\begin{tabular}{c|cc|ccc}
\toprule
\multirow{2}{*}{\textbf{Model Variants}} & \multicolumn{2}{c|}{\textbf{Link Prediction}} & \multicolumn{3}{c}{\textbf{Node Clustering}} \\ \cline{2-6} 
        & \textbf{AUC}   & \textbf{AP}    & \textbf{ACC}    & \textbf{NMI}    & \textbf{ARI}    \\ \midrule
ROD-no-$\mathcal{L}_t$           & 94.0   & 94.7  & 61.4  & 48.3  & 37.6\\
ROD-no-$\mathcal{L}_r$           & 95.6  & 95.9  & 72.2  & 55.2  & 50.1\\
ROD-no-$\mathcal{L}_d$           & 95.2   & 95.4  & 72.6  & 55.8  & 50.8\\ \midrule
ROD                    & \textbf{96.0}   & \textbf{96.4}  & \textbf{74.9}  & \textbf{58.3}  & \textbf{54.0}\\
\bottomrule
\end{tabular}}}
\vspace{-1mm}
\end{table}
\vspace{-1mm}
\paragraph{\underline{Task-based Supervision $\mathcal{L}_t$}} Experiments on both the link prediction and node clustering tasks indicate that the task-specific loss contributes the most as the performance of "ROD-no-$\mathcal{L}_t$" is significantly lower than the other two variants of our model.
For node clustering, $\mathcal{L}_t$ pushes all the nodes towards the centroid of their pseudo cluster and away from the other centroids, which directly boosts the performance of the node clustering task. Similarly, $\mathcal{L}_t$ in link prediction has enforced the learning of the adjacent matrix, while the other two losses contribute indirectly to the final result.

\vspace{-1mm}
\paragraph{\underline{Reception-aware information preservation $\mathcal{L}_r$}} 
Compared with the distillation loss $\mathcal{L}_d$, $\mathcal{L}_r$ plays a more important role in the node clustering task while $\mathcal{L}_d$ matters more in the link prediction. With $\mathcal{L}_r$, the multi-scale reception-aware graph knowledge global information can be preserved for each individual student model, thus enhancing the diversity of the ensemble teacher.

\vspace{-1mm}
\paragraph{\underline{Reception-aware online distillation $\mathcal{L}_d$} }
The distilled knowledge is crucial to ROD in both link prediction and node clustering. Take the node clustering task as an example, removing the distillation loss $\mathcal{L}_d$ leads to large performance degradation of 2.3\% in ACC, 2.5\% in NMI, and 3.2\% in ARI. With the distilled teacher knowledge, each student model can get sufficient and reliable supervision under the graph sparsity, and thus it can generate high-quality node embeddings for the downstream tasks.
The above results verify the rationality and effectiveness of each supervision signal (i.e., knowledge) of \sys.

\begin{figure}[tpb]
    \centering
    \includegraphics[width=0.7\linewidth]{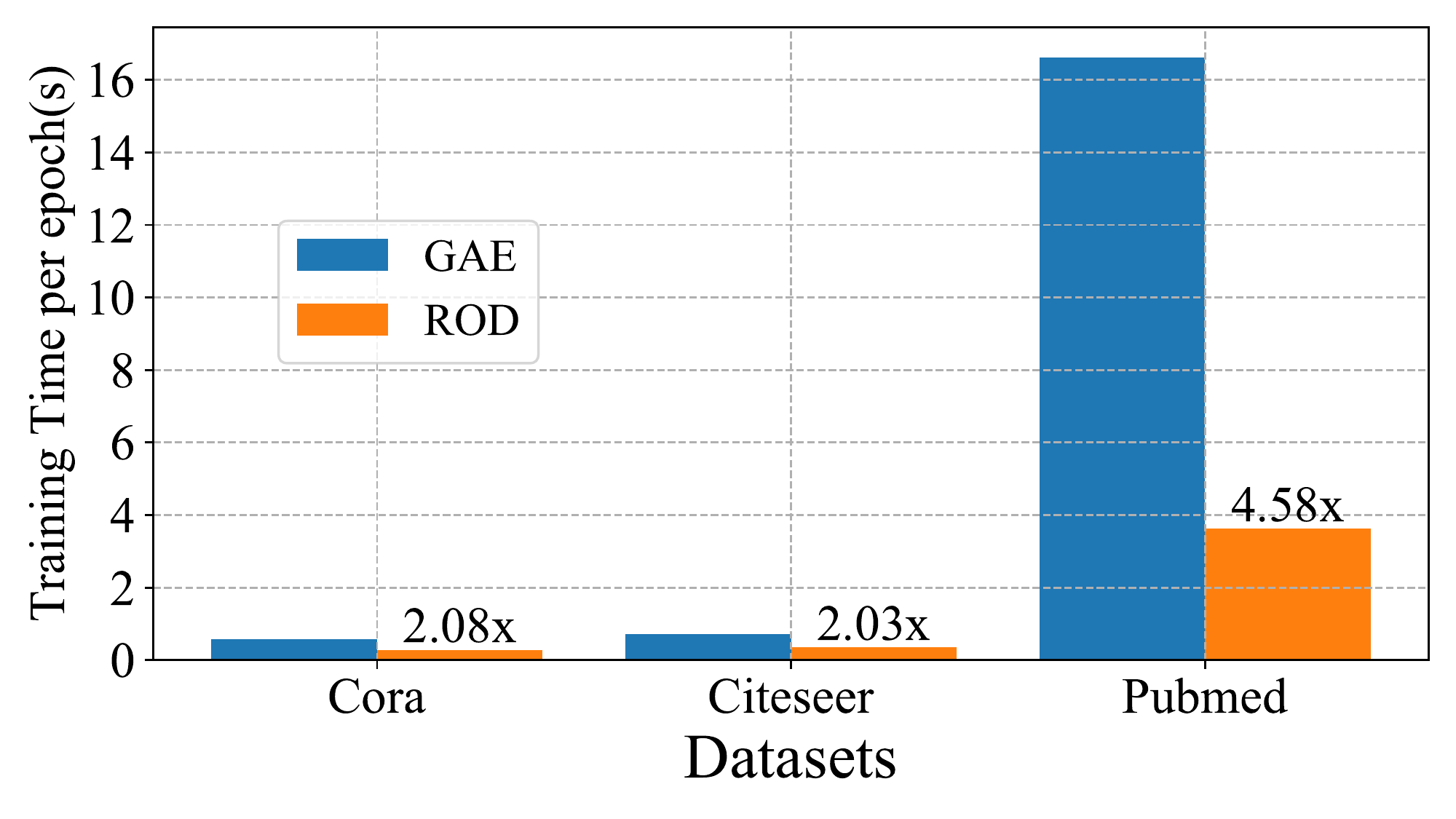}
    \vspace{-4mm}
    \caption{The compared training time and speedup on three citation networks.}
    \label{efficiency}
    \vspace{-4mm}
\end{figure}

\subsection{Efficiency Comparison} 
To validate the computational efficiency of the proposed ROD, we compare it with GAE in the link prediction task, which achieves the second-best performance.
In Figure ~\ref{efficiency}, we plot the efficiency of GAE and ROD in terms of training time on three graphs, with the propagation layers $k$=5. 
ROD includes the prepossessing time for the similarity matrix and sampling.

Figure ~\ref{efficiency} shows that ROD is faster across graphs, especially in the large PubMed graph. GAE takes reconstruction loss as the optimization objective, and the time complexity is $\mathcal{O}(N^{2})$ since it performs the calculation on every position of the adjacency matrix. With the importance sampling strategy, ROD only needs to calculate on a small part of the original given data rather than the entire graph, and it reduces the total time complexity from $\mathcal{O}(N^{2})$ to $\mathcal{O}(KM)$, where $K$ is the number of students models and $M$ is the edge number of the original graph.
Besides, compared with the dense layer in ROD, each graph convolution layer in GAE has an extra matrix multiplication operation on the adjacent matrix, making the training inefficient.

%% file: 5.appendix.tex
\appendix
\section{Outline}
The appendix is organized as follows:
\begin{description}
    \item[A.1] Datasets description.
    \item[A.2] More details about the compared baselines.
    \item[A.3] Experimental settings.
    \item[A.4] Experimental environment and reproduction instructions.
    \item[A.5] Motivation of label sparsity.
    \item[A.6] Motivation of decoupled GNNs.
\end{description}

\subsection{Dataset Description.}
\label{dataset_description}
\textbf{Cora}, \textbf{Citeseer}, and \textbf{Pubmed}\footnote{https://github.com/tkipf/gcn/tree/master/gcn/data} are three popular citation network datasets, and we follow the public training/validation/test split in GCN ~\cite{kipf2016semi}.
In these three networks, papers from different topics are considered as nodes and the edges are citations among the papers.  The node attributes are binary word vectors and class labels are the topics papers belong to.

\noindent\textbf{Amazon Computers} and \textbf{Amazon Photo} are segments of the Amazon co-purchase graph ~\cite{shchur2018pitfalls}. In these two networks, the nodes represent commodities on Amazon and the node features are binary words vectors encoded from commodity reviews. If two commodities are frequently bought together, there will be an edge between them. The class labels are naturally set to the commodity category.

\noindent\textbf{Coauthor CS} and \textbf{Coauthor Physics} are co-authorship graphs based on the Microsoft Academic Graph from the KDD Cup 2016 challenge\footnote{https://kddcup2016.azurewebsites.net/}. 
The authors are abstracted into nodes in these two networks, which are connected by an edge if they have co-authored a paper once. While the node features are composed of keywords for one’s papers. For class labels, they indicate the most active fields of research for each author. A pre-divided version of these datasets through the Deep Graph Library (DGL)\footnote{https://docs.dgl.ai/en/0.4.x/api/python/data.html\#coauthor-dataset} is used in our paper.

\noindent\textbf{Flickr} originates from NUS-wide~\footnote{http://lms.comp.nus.edu.sg/research/NUS-WIDE.html}. It contains a huge amount of images that are distributed to different types according to their descriptions and characteristics.

\noindent\textbf{Reddit} is a social network dataset derived from the community structure of numerous Reddit posts. It is a well-known inductive training dataset and the training/validation/test split in our experiment is the same as the split in GraphSAGE~\cite{hamilton2017inductive}. The public version of Reddit and Flickr provided by GraphSAINT\footnote{https://github.com/GraphSAINT/GraphSAINT} is used in our paper.

\subsection{Compared Baselines.}
The main characteristic of all baselines are listed below:
\label{baseline_methods}
\begin{itemize}
    \item Graph Convolutional Network(GCN) ~\cite{kipf2016semi}: GCN is a novel and efficient method for semi-supervised classification on graph-structured data.
    \item Graph Attention Networks(GAT) ~\cite{velivckovic2017graph}: GAT leverages masked self-attention layers to specify different weights to different nodes in a neighborhood, thus better represent graph information.
    \item Jumping Knowledge Networks(JK-Net) ~\cite{xu2018representation}: JK-Net is a flexible network embedding method that could gather different neighborhood ranges to enable better structure-aware representation. 
    \item APPNP ~\cite{klicpera2018predict}: APPNP uses the relationship between graph convolution networks (GCN) and PageRank to derive improved node representations.
    \item Adaptive Propagation Graph Convolution Network (AP-GCN) ~\cite{spinelli2020adaptive}: AP-GCN uses a halting unit to decide a receptive range of a given node. 
    \item GraphSAGE ~\cite{hamilton2017inductive}: GraphSAGE is an inductive framework that leverages node attribute information to efficiently generate representations on previously unseen data.
    \item FastGCN ~\cite{chen2018fastgcn}: FastGCN interprets graph convolutions as integral transforms of embedding functions under probability measures.
    \item Cluster-GCN ~\cite{chiang2019cluster}: Cluster-GCN is a novel GCN algorithm that is suitable for SGD-based training by exploiting the graph clustering structure.
    \item GraphSAINT ~\cite{zeng2019graphsaint}: GraphSAINT constructs mini-batches by sampling the training graph, rather than the nodes or edges across GCN layers.
    \item Self-training ~\cite{li2018deeper}: Self-training enlarges the training set during the training process by selecting the most confident predictions for each class according to softmax outputs.
    \item Multi-Stage Self-Supervised (M3S) ~\cite{sun2020multi}: Based on a GCN with multi-scale signals, M3S introduces a special aligning mechanism into the conventional self-training method.
    \item Graph Autoencoder (GAE) and Variational Graph Autoencoder (VGAE) ~\cite{kipf2016variational}: GAE and VGAE are methods for unsupervised graph-structured representations learning based on Graph Autoencoder and Variational Graph Autoencoder. 
    \item Adversarially Regularized Graph Autoencoder (ARGA) and Adversarially Regularized Variational Graph Autoencoder (ARVGA) ~\cite{pan2018adversarially}: ARGA and ARVGA use the adversarial learning strategy, and the generated embeddings are forced to match a prior distribution. 
    \item Marginalized Graph Autoencoder (MGAE) ~\cite{wang2017mgae}: MGAE is a denoising marginalized graph autoencoder that reconstructs the attribute matrix, taking both graph structure and node attributes into consideration.
    \item Adaptive Graph Convolution (AGC) ~\cite{zhang2019attributed}: AGC is a graph convolution method for attributed graph clustering that adaptively select high-order graph information to capture global cluster structure.
    \item Deep Attentional Embedded Graph Clustering (DAEGC) ~\cite{wang2019attributed}: DAEGC uses side information to generate graph embedding for clustering and trains in a self-training way.
    \item Structural Deep Clustering Network (SDCN) ~\cite{bo2020structural}: SDCN uses an additional autoencoder to obtain node attribute information, and a self-training method is adapted.
    \item Adaptive Graph Encoder (AGE) ~\cite{cui2020adaptive}: AGE is a flexible embedding framework with a Laplacian smoothing filter and an adaptive encoder for high-quality node embedding.
    \item Simplified GCN (SGC) ~\cite{wu2019simplifying}: SGC simplifies GCN by removing nonlinearities and collapsing weight matrices between consecutive layers. 
    \item Scalable Inception Graph Neural Networks (SIGN) ~\cite{rossi2020sign}: SIGN is an efficient and scalable graph embedding method that sidesteps graph sampling in GCN and uses different local graph operators to support different tasks. 
    \item Spectral Clustering (SC) ~\cite{ng2001spectral}: SC is an approach based on matrix factorization, generating the vertex representation with the smallest d eigenvectors of the normalized Laplacian matrix of the graph.
    \item DeepWalk (DW) ~\cite{perozzi2014deepwalk}: DW is a skip-gram based model that learns the graph embedding with truncated random walks.
\end{itemize}

\subsection{Experimental Settings.}
\label{hyperparameter_setting}
\paragraph{Link Prediction Task}
In the link prediction task, the validation and test sets contain 5\% and 10\% of citation links, respectively. 
For LightGCN, SGC and SIGN, hyperparameters were tuned on ROC and AUC scores using a validation set. 
For GAE and VGAE, the hidden size is set to 32 and the learning rate is set to 0.01. For DW, the number of walks per vertex in DeepWalk is set to 10 with walk length 30, and the window size is optimized to 10. For other baselines, we use the implementations provided by the authors, and we set the hyperparameters to the default values mentioned in their papers. 
For ROD, we train for 400 epochs, and we keep the same parameters in both Cora and Citeseer. We set the number of hidden units 1024, the learning rate of 0.001, and k to 4. We set $\alpha$ 0.2 and $\beta$ 0.1 for both.
To validate the effectiveness of ROD, we train our model on different datasets and make an end-to-end comparison on the link prediction task. 

\paragraph{Node Clustering Task}
The hyperparameters for the baseline methods are all set as recommended in their original paper.
While for ROD, the dimension of the embedding is set to 64, and k is set to 4.
The learning rate is 0.01 for Cora and 0.001 for Citeseer and Pubmed.
We set both $\alpha$ and $\beta$ to 0.1 for Cora and both to 10 for Citeseer and Pubmed.
For end-to-end methods like DAEGC and our model ROD, the node clustering results are obtained directly from the training process.
While for others, we run K-Means on the learned embedding to get the node clustering results.
All the methods are trained for 200 epochs using the Adam optimizer.

\paragraph{Node Classification Task}
All the baselines follow the same hyper-parameter in their original paper.
We train our models using Adam optimizer with a learning rate of 0.02 for citation networks, 0.05 for two co-authorship graphs, 0.01 for the two amazon co-purchase graphs and Flickr, and 0.005 for Reddit. The regularization factor is 5e-4 for the citation networks and to 1e-5 for other datasets. 
We set both $\alpha$ and $\beta$ to 0.1 for three citation networks, and set $\alpha$ 0.1 and $\beta$ 0.3 for other datasets.
The dropout is applied to all feature vectors with rates of 0.8 for the citation networks and 0.5 for other datasets.
We adopt one dense layer as the feature encoder and the hidden size is 128. For each student model, it uses an extra dense layer to learn its individual parameters.
Besides, the training budget is 200 epochs for all datasets.
To eliminate random factors, we run each method 20 times and report the mean and variance of the corresponding test accuracy. 

\subsection{ Experiment Environment and Reproduction Instructions}
We run our experiments on a machine with 14 Intel(R) Xeon(R) CPUs (Gold 5120 @ 2.20GHz) and four NVIDIA TITAN RTX GPUs.
The code is written in Python 3.6. We use Pytorch 1.7.1 on CUDA 10.1 to train the model on GPU.
Our code is publicly available on https://github.com/zwt233/ROD.

\subsection{Motivation of Label Sparsity}
\begin{figure}[tpb]
    \centering
    \includegraphics[width=0.8\linewidth]{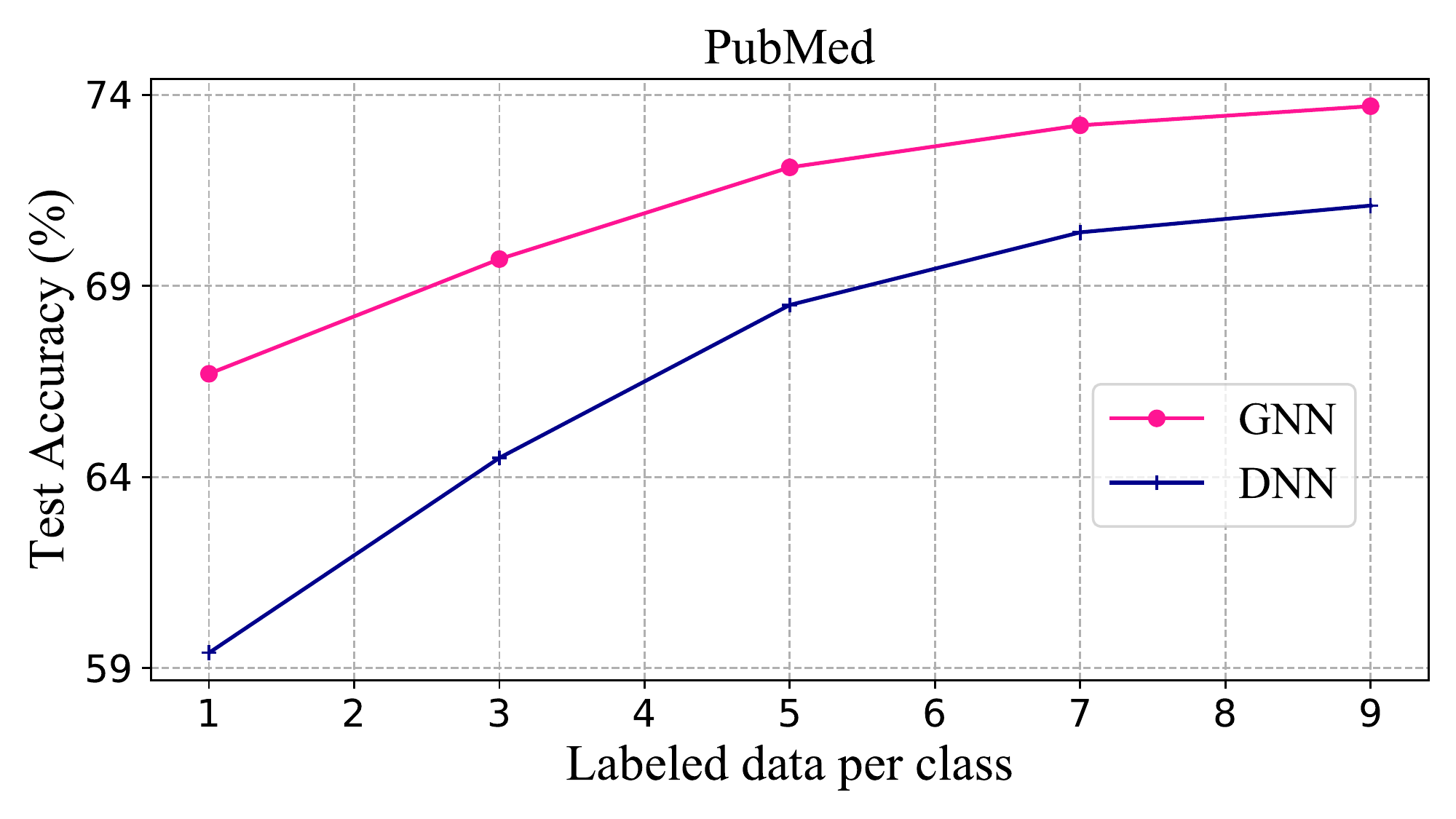}
    \vspace{-4mm}
    \caption{Performance of GCN and DNN using different number of labeled data on PubMed dataset.}
    \label{spar}
\end{figure}

Figure~\ref{spar} shows the performance gain between GNN and DNN (both two layers, and the GNN model is APPNP) is significantly large in the low label setting (e.g., less than 3 labeled nodes per class). 
Because adding the labeled data to GNN will further incorporate more unlabeled nodes into the model training, as compared to DNN.
Besides, the performance gain gradually decreased as the label and feature information has already propagated a sufficient number of unlabeled nodes.
The large marginal gain in low label setting indicates a high-performance improvement potential of GNN under label sparsity.

\begin{figure}[!htpb]
    \centering
    \includegraphics[width=\linewidth]{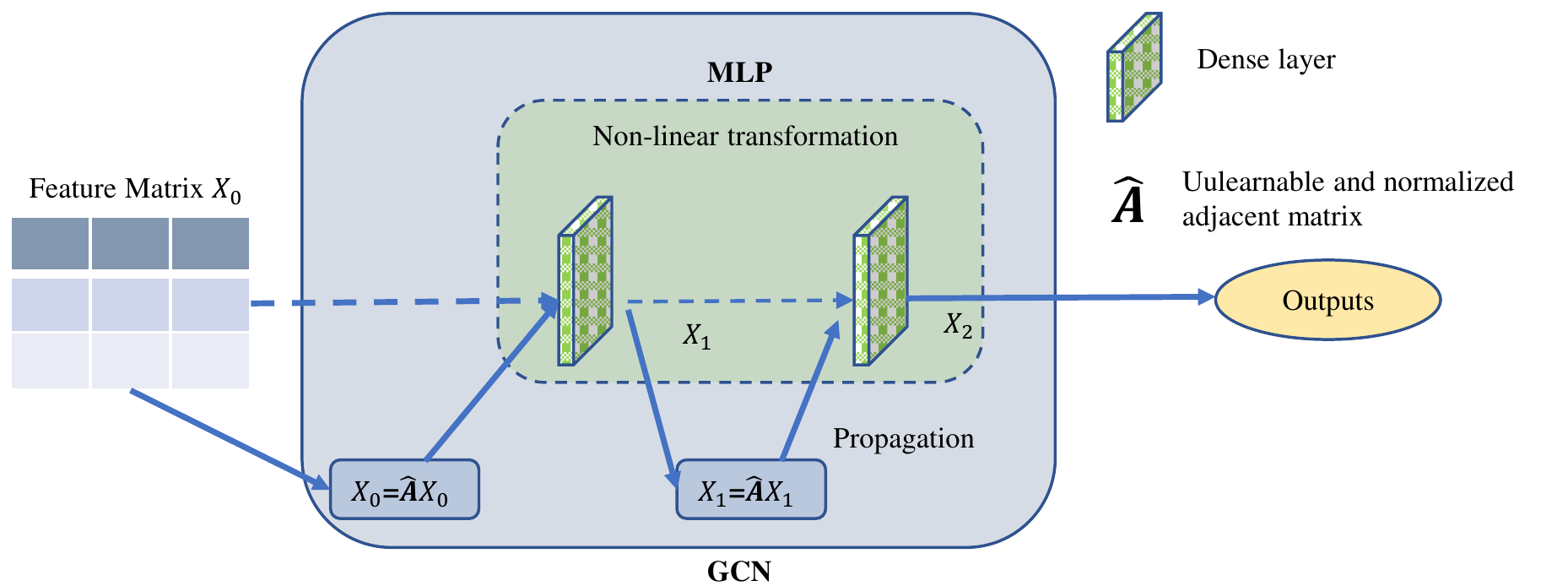}
    \vspace{-2mm}
    \caption{The framework of GCN}
    \vspace{-4mm}
    \label{GCN-fra}
\end{figure}

\subsection{Motivation of Decoupled GNNs}

Figure~\ref{GCN-fra} shows the framework of a two-layers GCN from the decouple view. The graph convolution layer in GCN is decoupled into nonlinear transformation and propagation and GCN will degenerate to MLP if we remove all the propagation process.
The graph convolution layer in GCN will first propagate and enhance the node representation with the neighbour nodes and then feed the representation to the non-linear transformation process. 
Lots of works observe that the entanglement of nonlinear transformation and propagation will harm both the performance and robustness, which motivates us to propose our decoupled GNN.